\RequirePackage{fix-cm}
\documentclass[twocolumn]{svjour3}          % twocolumn
\smartqed  % flush right qed marks, e.g. at end of proof

\usepackage{natbib}
\usepackage{amssymb}
\usepackage{graphicx}
\usepackage{url}
\usepackage{multirow}
\usepackage{mathtools}
\usepackage{balance}
\usepackage{hhline}
\usepackage{makecell}
\usepackage{epsfig}
\usepackage{subfigure}
\usepackage[dvipsnames]{xcolor}
\usepackage[pagebackref=true,breaklinks=true,letterpaper=true,colorlinks,bookmarks=false]{hyperref}
\usepackage[nice]{nicefrac}

\newcolumntype{Y}{>{\centering\arraybackslash}X}
\newcolumntype{M}[1]{>{\centering\arraybackslash}m{#1}}
\newcolumntype{?}{!{\vrule width 1pt}}

\makeatletter
\newcommand*\bigcdot{\mathpalette\bigcdot@{.5}}
\newcommand*\bigcdot@[2]{\mathbin{\vcenter{\hbox{\scalebox{#2}{$\m@th#1\bullet$}}}}}

\newcommand{\ssymbol}[1]{^{\@fnsymbol{#1}}}
\def\@fnsymbol#1{\ensuremath{\ifcase#1\or *\or \dagger\or \ddagger\or
   \mathsection\or \mathparagraph\or \|\or **\or \dagger\dagger
   \or \ddagger\ddagger \else\@ctrerr\fi}}
   
\DeclareMathOperator{\EX}{\mathbb{E}}% expected value

\begin{document}

\emergencystretch 1.5em

\title{DnS: Distill-and-Select for Efficient and Accurate Video Indexing and Retrieval%\thanks{Grants or other notes
%about the article that should go on the front page should be
%placed here. General acknowledgments should be placed at the end of the article.}
}
% \subtitle{Do you have a subtitle?\\ If so, write it here}

%\titlerunning{Short form of title}        % if too long for running head

\author{Giorgos Kordopatis-Zilos \and
        Christos Tzelepis \and
		Symeon Papadopoulos \and
        Ioannis Kompatsiaris \and
        Ioannis Patras
}

%\authorrunning{Short form of author list} % if too long for running head

% \institute{Giorgos Kordopatis-Zilos \at
%               Information Technologies Institute, Centre for Research and Technology Hellas, Thessaloniki, Greece, and Queen Mary University of London, Mile End road, E1 4NS London \\
%               \email{georgekordopatis@iti.gr}
%             \and
%             Christos Tzelepis \and Ioannis Patras \at
%               Queen Mary University of London, Mile End road, E1 4NS London \\
%               \email{\{c.tzelepis, i.patras\}@qmul.ac.uk}
%             \and
%             Symeon Papadopoulos \and Ioannis Kompatsiaris \at
%               Information Technologies Institute, Centre for Research and Technology Hellas, Thessaloniki, Greece \\
%               \email{\{papadop, ikom\}@iti.gr}
% }

\institute{Giorgos Kordopatis-Zilos -- ITI-CERTH, QMUL \at
              \email{georgekordopatis@iti.gr}
            \and
            Christos Tzelepis \and Ioannis Patras -- QMUL \at
              \email{\{c.tzelepis, i.patras\}@qmul.ac.uk}
            \and
            Symeon Papadopoulos \and Ioannis Kompatsiaris -- ITI-CERTH \at
              \email{\{papadop, ikom\}@iti.gr}
}

\date{Received: date / Accepted: date}
% The correct dates will be entered by the editor

\maketitle

\begin{figure}[t]
    \centering
    \includegraphics[width=\linewidth]{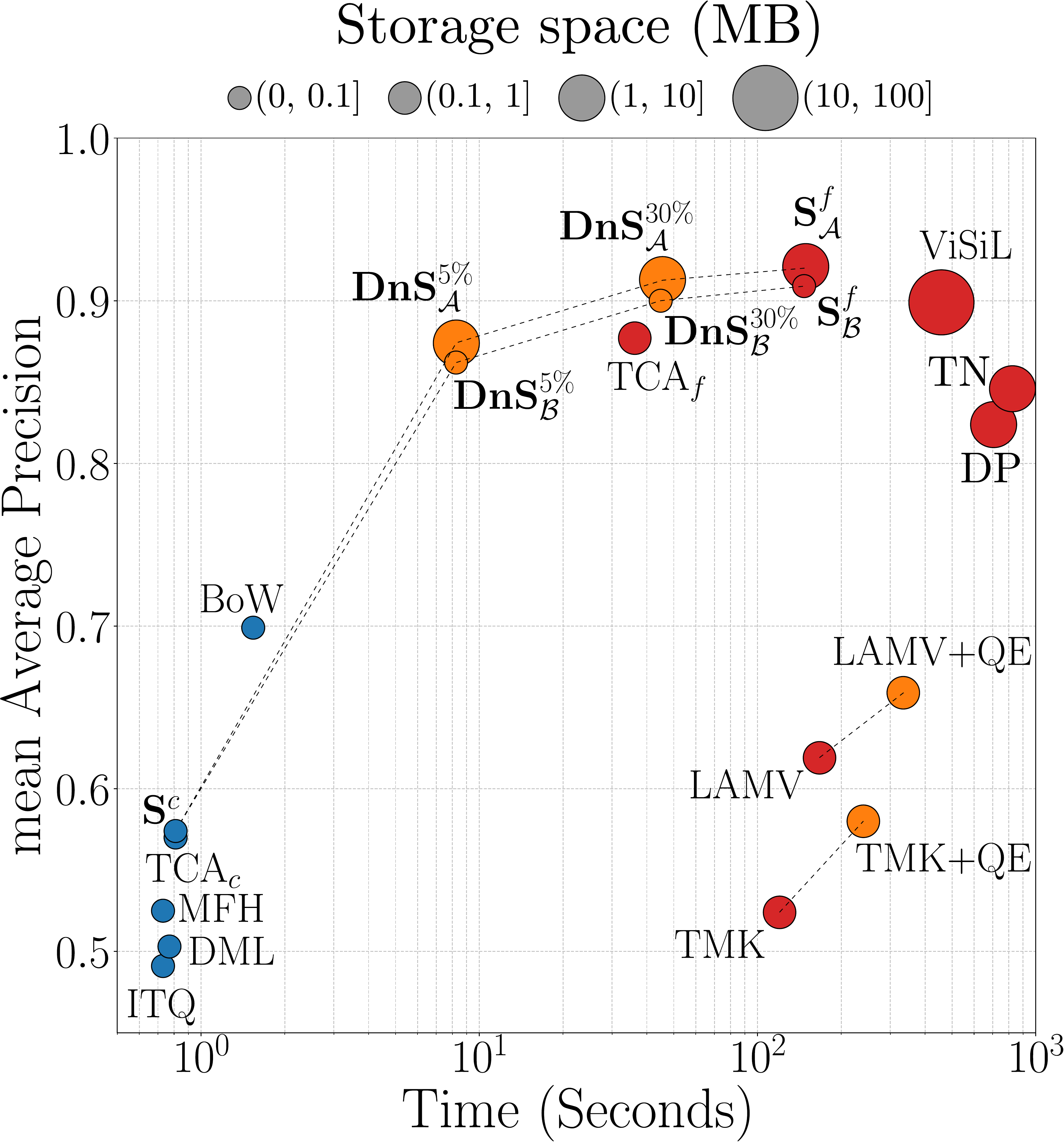}
    \caption{Performance of our proposed \textbf{DnS} framework and its variants for several dataset percentages sent for re-ranking (denoted in \textbf{bold}) evaluated on the DSVR task of FIVR-200K in terms of mAP, computational time per query in seconds, and storage space per video in megabytes (MB), in comparison to state-of-the-art methods. Coarse-grained methods are in \textcolor[HTML]{1f77b4}{blue}, fine-grained in \textcolor[HTML]{d62728}{red}, and re-ranking in \textcolor[HTML]{ff7f0e}{orange}.}
    \label{fig:performance}
\end{figure}

\begin{abstract}
In this paper, we address the problem of high performance and computationally efficient content-based video retrieval in large-scale datasets. Current methods typically propose either: (i) fine-grained approaches employing spatio-temporal representations and similarity calculations, achieving high performance at a high computational cost or (ii) coarse-grained approaches representing/indexing videos as global vectors, where the spatio-temporal structure is lost, providing low performance but also having low computational cost. In this work, we propose a Knowledge Distillation framework, called Distill-and-Select (DnS), that starting from a well-performing fine-grained Teacher Network learns: a) Student Networks at different retrieval performance and computational efficiency trade-offs and b) a Selector Network that at test time rapidly directs samples to the appropriate student to maintain both high retrieval performance and high computational efficiency. We train several students with different architectures and arrive at different trade-offs of performance and efficiency, i.e., speed and storage requirements, including fine-grained students that store/index videos using binary representations. Importantly, the proposed scheme allows Knowledge Distillation in large, unlabelled datasets -- this leads to good students. We evaluate DnS on five public datasets on three different video retrieval tasks and demonstrate a) that our students achieve state-of-the-art performance in several cases and b) that the DnS framework provides an excellent trade-off between retrieval performance, computational speed, and storage space. In specific configurations, the proposed method achieves similar mAP with the teacher but is 20 times faster and requires 240 times less storage space. The collected dataset and implementation are publicly available: \url{https://github.com/mever-team/distill-and-select}.
\end{abstract}

\section{Introduction}\label{sec:introduction}
    
    \begin{figure*}[t]
        \centering
        \includegraphics[width=\linewidth]{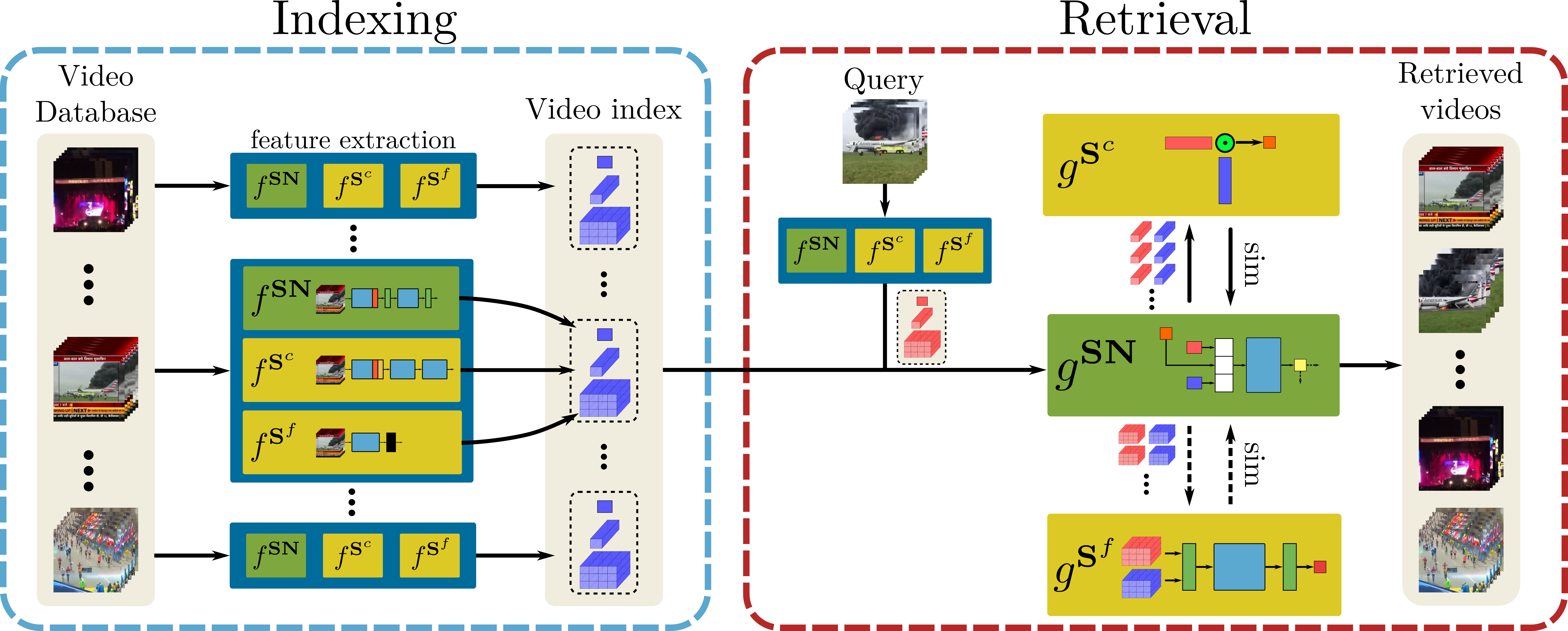}
        \caption{Overview of the proposed framework. It consists of three networks: a coarse-grained student $\textbf{S}^c$, a fine-grained student $\textbf{S}^f$, and a selector network $\textbf{SN}$. Processing is split into two phases, \textit{Indexing} and \textit{Retrieval}. During indexing (blue box), given a video database, three representations needed by our networks are extracted and stored in a video index, i.e., for each video, we extract a 3D tensor, a 1D vector, and a scalar that captures video self-similarity. During retrieval (red box), given a query video, we extract its features, which, along with the indexed ones, are processed by the $\textbf{SN}$. It first sends all the 1D vectors of query-target pairs to $\textbf{S}^c$ for an initial similarity calculation. Then, based on the calculated similarity and the self-similarity of the videos, the selector network judges which query-target pairs have to be re-ranked with the $\textbf{S}^f$, using the 3D video tensors. Straight lines indicate continuous flow, i.e., all videos/video pairs are processed, whereas dashed lines indicate conditional flow, i.e., only a number of selected videos/video pairs are processed. Our students are trained with Knowledge Distillation based on a fine-grained teacher network, and the selector network is trained based on the similarity difference between the two students.}
        \label{fig:overview}
    \end{figure*}
    
    % === Introductory paragraph -- 
    Due to the popularity of Internet-based video sharing services, the volume of video content on the Web has reached unprecedented scales. For instance, YouTube reports that more than 500 hours of content are uploaded every minute\footnote{\url{https://www.youtube.com/yt/about/press/}, accessed June 2021}. This puts considerable challenges for all video analysis problems, such as video classification, action recognition, and video retrieval, that need to achieve high performance at low computational and storage requirements in order to deal with the large scale of the data. The problem is particularly hard in the case of content-based video retrieval, where, given a query video, one needs to calculate its similarity with all videos in a database to retrieve and rank the videos based on relevance. In such scenario, this requires efficient indexing, i.e., storage of the representations extracted from the videos in the dataset, and fast calculations of the similarity between pairs of them.

    % === Coarse- and fine-grained methods ===
    % AKIS: Given that there is a related work section, I think that this may be too detailed at this point
    Depending on whether the spatio-temporal structure of videos is stored/indexed and subsequently taken into consideration during similarity calculation, research efforts fall into two broad categories, namely \textit{coarse-} and \textit{fine-grained} approaches. \textit{Coarse-grained} approaches address this problem by aggregating frame-level features into single video-level vector representations (that are estimated and stored at indexing time) and then calculating the similarity between them by using a simple function such as the dot-product or the Euclidean distance (at retrieval time). The video-level representations can be global vectors \citep{gao2017,kordopatis2017b,lee2020}, hash codes \citep{song2011,song2018,yuan2020}, Bag-of-Words (BoW) \citep{cai2011,kordopatis2017a,liao2018}, or concept annotations \citep{markatopoulou2017,markatopoulou2018,liang2020}. These methods have very low storage requirements, allow rapid similarity estimation at query-time, but they exhibit low retrieval performance, since they disregard the spatial and temporal structure of the videos and are therefore vulnerable to clutter and irrelevant content. On the other hand, \textit{fine-grained} approaches extract (and store at indexing time) and use in the similarity calculation (at retrieval time) representations that respect the spatio-temporal structure of the original video, i.e., they have a temporal or a spatio-temporal dimension/index. Typically, such methods consider the sequence of frames in the similarity calculation and align them, e.g., by using Dynamic Programming~\citep{chou2015,liu2017}, Temporal Networks~\citep{tan2009,jiang2016}, or Hough Voting~\citep{douze2010,jiang2014}; or consider spatio-temporal video representation and matching based on Recurrent Neural Networks (RNN)~\citep{feng2018,bishay2019}, Transformer-based architectures~\citep{shao2021}, or in the Fourier domain~\citep{poullot2015,baraldi2018}. These approaches achieve high retrieval performance but at considerable computation and storage cost. 
    
    % === Re-ranking methods (combination of fine- and coarse-grained methods) ===
    In an attempt to exploit the merits of both fine- and coarse-grained methods, some works tried to utilize them in a single framework~\citep{wu2007,chou2015,liang2020}, leading to methods that offer a trade-off between computational efficiency and retrieval performance. Typically, these approaches first rank videos based on a coarse-grained method, in order to filter the videos with similarity lower than a predefined threshold, and then re-rank the remaining ones based on the similarity calculated from a computationally expensive fine-grained method. However, setting the threshold is by no means a trivial task. In addition, in those approaches, both coarse- and fine-grained components are typically built based on hand-crafted features with traditional aggregations (e.g., BoW) and heuristic/non-learnable approaches for similarity calculation -- this results in sub-optimal performance. We will be referring to such approaches as \textit{re-ranking} methods.
    
    Fig. \ref{fig:performance} illustrates the retrieval performance, time per query, and storage space per video of several methods from the previous categories. Fine-grained approaches achieve the best results but with a significant allocation of resources. On the other hand, coarse-grained approaches are very lightweight but with considerably lower retrieval performance. Finally, the proposed re-ranking method provides a good trade-off between accuracy and efficiency, achieving very competitive performance with low time and storage requirements.
    
    % === Knowledge Distillation ===
    Knowledge Distillation is a methodology in which a student network is being trained so as to approximate the output of a teacher network, either in the labelled dataset in which the teacher was trained, or in other, potentially larger unlabelled ones. Depending on the student's architecture and the size of the dataset, different efficiency-performance trade-offs can be reached. These methods have been extensively used in the domain of image recognition \citep{yalniz2019,touvron2020,xie2020}; however, in the domain of video analysis, they are limited to video classification methods \citep{bhardwaj2019,garcia2018,crasto2019,stroud2020}, typically performing distillation at feature level across different modalities. Those methods typically distill the features of a stream of the network operating in a (computationally) expensive modality (e.g., optical flow field, or depth) into the features of a cheaper modality (e.g., RGB images) so that only the latter need to be stored/extracted and processed at test time. This approach does not scale well on large datasets, as it requires storage or re-estimation of the intermediate features. Furthermore, current works arrive at fixed trade-offs of performance and computational/storage efficiency.
    
    % === Contributions of this work ===
    % AKIS: I would perhaps move some of the details from the following paragraph to the bullet-contributions.
    In this work, we propose to address the problem of high retrieval performance and computationally efficient content-based video retrieval in large-scale datasets. The proposed method builds on the framework of Knowledge Distillation, and starting from a well-performing, high-accuracy-high-complexity teacher, namely a fine-grained video similarity learning method (ViSiL)~\citep{kordopatis2019b}, trains a) both fine-grained and coarse-grained student networks on a large-scale unlabelled dataset and b) a selection mechanism, i.e., a learnable re-ranking module, that decides whether the similarity estimated by the coarse-grained student is accurate enough, or whether the fine-grained student needs to be invoked. By contrast to other re-ranking methods that use a threshold on the similarity estimated by the fast network (the coarse-grained student in our case), our selection mechanism is a trainable, lightweight neural network. All networks are trained so as to extract representations that are stored/indexed, so that each video in the database is indexed by the fine-grained spatio-temporal representation (3D tensor), its global, vector-based representation (1D vector), and a scalar self-similarity measure that is extracted by the feature extractor of the selector network, and can be seen as a measure of the complexity of the videos in question. The latter is expected to be informative of how accurate the coarse-grained, video-level similarity is, and together with the similarity rapidly estimated by the coarse-grained representations, is used as input to the selector. We note that, by contrast to other Knowledge Distillation methods in videos that address classification problems and typically perform distillation at intermediate features, the students are trained on a similarity measure provided by the teacher -- this allows training on large scale datasets as intermediate features of the networks do not need to be stored, or estimated multiple times. Due to the ability to train on large unlabeled datasets, more complex models, i.e., with more trainable parameters, can be employed leading to even better performance than the original teacher network. An overview of the proposed framework is illustrated in Fig.~\ref{fig:overview}.
    
    The main contributions of this paper can be summarized as follows:
    \begin{itemize}
        \item We build a re-ranking framework based on a Knowledge Distillation scheme and a Selection Mechanism that allows for training our student and selector networks using large unlabelled datasets. We employ a teacher network that is very accurate but needs a lot of computational resources to train several student networks and the selector networks, and use them to achieve different performance-efficiency trade-offs.
        \item We propose a selection mechanism that, given a pair of a fine- and a coarse-grained student, learns whether the similarity estimated by the fast, coarse-grained student is accurate enough, or whether the slow, fine-grained student needs to be invoked. To the best of our knowledge, we are the first to propose such a trainable selection scheme based on video similarity.
        \item We propose two fine-grained and one coarse-grained student architectures. We develop: (i) a fine-grained attention student, using a more complex attention scheme than the teacher's, (ii) a fine-grained binarization student that extracts binarized features for the similarity calculation, and (iii) a course-grained attention student that exploits region-level information, and the intra- and inter-video relation of frames for the aggregation.
        \item We evaluate the proposed method on five publicly available datasets and compare it with several state-of-the-art methods. Our fine-grained student achieves state-of-the-art performance on two out of four datasets, and our DnS approach retains competitive performance with more than 20 times faster retrieval per query and 99\% lower storage requirements compared to the teacher.
    \end{itemize}
    
    The remainder of the paper is organised as follows. In Sect.~\ref{sec:related_work}, the related literature is discussed. In Sect.~\ref{sec:proposed_method}, the proposed method is presented in detail. In Sect.~\ref{sec:evaluation_setup}, the datasets and implementation are presented. In Sect.~\ref{sec:experiments}, the results and ablation studies are reported. In Sect.~\ref{sec:conclusion}, we draw our conclusions.

%%%%%%%%%%%%%%%%%%%%%%%%%%%%%%%%%%%%%%%%%%%%%%%%%%%%%%%%%%%%%%%%%%%%%%%%%%%%%%%%
%%                                                                            %%
%%                              [Related Work]                                %%
%%                                                                            %%
%%%%%%%%%%%%%%%%%%%%%%%%%%%%%%%%%%%%%%%%%%%%%%%%%%%%%%%%%%%%%%%%%%%%%%%%%%%%%%%%
\section{Related Work}\label{sec:related_work}
    This section gives an overview of some of the fundamental works that have contributed to content-based video retrieval and knowledge distillation.
    
    \subsection{Video retrieval}
        The video retrieval methods can be roughly classified, based on the video representations and similarity calculation processes employed, in three categories: coarse-grained, fine-grained, and re-ranking approaches.
        
        \subsubsection{Coarse-grained approaches}
        Coarse-grained approaches represent videos with a global video-level signature, such as an aggregated feature vector or a binary hash code, and use a single operation for similarity calculation, such as a dot product. A straightforward approach is the extraction of global vectors as video representations combined with the dot product for similarity calculation. Early works~\citep{wu2007,huang2010} extracted hand-crafted features from video frames, i.e., color histograms, and aggregated them to a global vector. More recent works~\citep{gao2017,kordopatis2017b,lee2018,lee2020} rely on CNN features combined with aggregation methods. Also, other works~\citep{cai2011,kordopatis2017a} aggregate video content to Bag-of-Words (BoW) representation~\citep{sivic2003} by mapping frames to visual words and extracting global representations with tf-idf weighting. Another popular direction is the generation of hash codes for the entire videos combined with Hamming distance~\citep{song2011,song2018,liong2017,yuan2020}. Typically, the hashing is performed via a network trained to preserve relations between videos. Coarse-grained methods provide very efficient retrieval covering the scalability needs of web-scale applications; however, their retrieval performance is limited, typically outperformed by the fine-grained approaches.
        
        \subsubsection{Fine-grained approaches}
        Fine-grained approaches extract video representations, ranging from video-level to region-level, and calculate similarity by considering spatio-temporal relations between videos based on several operations, e.g., a dot product followed by a max operation. \cite{tan2009} proposed a graph-based Temporal Network (TN) structure, used for the detection of the longest shared path between two compared videos, which has also been combined with frame-level deep learning networks~\citep{jiang2016,wang2017}. Additionally, other approaches employ Temporal Hough Voting~\citep{douze2010} to align matched frames by means of a temporal Hough transform. Another solution is based on Dynamic Programming (DP)~\citep{chou2015}, where the similarity matrix between all frame pairs is calculated, and then the diagonal blocks with the largest similarity are extracted. Another direction is to generate spatio-temporal representations with the Fourier transform in a way that accounts for the temporal structure of video similarity~\citep{poullot2015, baraldi2018}. Finally, some recent works rely on attention-based schemes to learn video comparison and aggregation by training either attentional RNN architectures~\citep{feng2018,bishay2019}, transformer-based networks for temporal aggregation~\citep{shao2021}, or multi-attentional networks that extract multiple video representations~\citep{wang2021}. Fine-grained methods achieve high retrieval performance; however, they do not scale well to massive datasets due to their high computational and storage requirements.
    
        \subsubsection{Video Re-ranking}
        Re-ranking is a common practice in retrieval systems. In the video domain, researchers have employed it to combine methods from the two aforementioned categories (i.e., coarse- and fine-grained) to overcome their bottleneck and achieve efficient and accurate retrieval~\citep{wu2007,douze2010,chou2015,yang2019,liang2020}. Typical methods deploy a coarse-grained method as an indexing scheme to quickly rank and filter videos, e.g., using global vectors~\citep{wu2007} or BoW representations~\citep{chou2015,liang2020}. Then, a fine-grained algorithm, such as DP~\citep{chou2015}, Hough Voting~\citep{douze2010} or frame-level matching~\citep{wu2007}, is applied on the videos that exceed a similarity threshold in order to refine the similarity calculation. Another re-ranking approach employed for video retrieval is Query Expansion (QE)~\citep{chum2007}. It is a two-stage retrieval process where, after the first stage, the query features are re-calculated based on the most similar videos retrieved, and the query process is executed again with the new query representation. This has been successfully employed with both coarse-grained~\citep{douze2013,gao2017,zhao2019} and fine-grained~\citep{poullot2015,baraldi2018} approaches. Also, an attention-based trainable QE scheme has been proposed in~\citep{gordo2020} for image retrieval. However, even though the retrieval performance is improved with QE, the total computational time needed for retrieval is doubled as the query process is applied twice.
    
    \subsection{Knowledge distillation}
        Knowledge Distillation~\citep{hinton2015} is a training scheme that involves two networks, a teacher network that is usually trained with supervision and a student network that leverages teacher's predictions for improved performance. A thorough review of the field is provided in~\citep{gou2021}. Knowledge Distillation has been employed on various computer vision problems, i.e., image classification~\citep{yalniz2019,touvron2020,xie2020}, object detection~\citep{li2017,shmelkov2017,deng2019}, metric learning~\citep{park2019,peng2019}, action recognition~\citep{garcia2018,thoker2019,stroud2020}, video classification~\citep{zhang2018,bhardwaj2019}, video captioning~\citep{pan2020,zhang2020}, and representation learning~\citep{tavakolian2019,piergiovanni2020}. 
    
        Relevant works in the field of Knowledge Distillation distill knowledge based on the relations between data samples \citep{park2019,tung2019,liu2019,lassance2020,peng2019}. Student networks are trained based on the distances between samples calculated by a teacher network~\citep{park2019}, the pairwise similarity matrix between samples within-batch~\citep{tung2019}, or by distilling graphs constructed based on the relations of the samples, using the sample representations as vertices and their distance as the edges to build an adjacency matrix~\citep{liu2019,lassance2020}.

        In the video domain, several approaches have been proposed for the improvement of the computational efficiency of the networks~\citep{bhardwaj2019,zhang2018,garcia2018}. Some works~\citep{bhardwaj2019} proposed a Knowledge Distillation setup for video classification where the student uses only a fraction of the frames processed by the teacher, or multiple teachers are employed to construct a graph based on their relations, and then a smaller student network is trained~\citep{zhang2018}. Also, a popular direction is to build methods for distillation from different modalities and learn with privileged information to increase the performance of a single network, i.e., using depth images~\citep{garcia2018}, optical flow~\citep{crasto2019,stroud2020,piergiovanni2020}, or multiple modalities~\citep{luo2018,piergiovanni2020}. In video retrieval, Knowledge Distillation has been employed for feature representation learning on frame-level using the evaluation datasets~\citep{liang2019}.
    
    \subsection{Comparison to previous approaches}
        In this section, we draw comparisons of the proposed approach to the related works from the literature with respect to the claimed novelties.
        
        \textbf{Proposed Framework:} There is no similar prior work in the video domain that builds a re-ranking framework based on Knowledge Distillation and a trainable Selection Mechanism based on which the re-ranking process is performed. Other works~\citep{chou2015,yang2019,liang2020} rely on outdated hand-crafted methods using simple re-ranking approaches based on similarity thresholding, the selection of which is a non-trivial task. By contrast, in this work, a framework is proposed that starts from an accurate but heavy-weight teacher to train a) both a fine-grained and coarse-grained student network on a large unlabelled dataset and b) a selection mechanism, i.e., a learnable module based on which the re-ranking process is performed.
        
        \textbf{Knowledge Distillation:} To the best of our knowledge, there is no prior work in the video domain that trains a pairwise function that measures video similarity with distillation. Works that use a similar loss function for distillation are~\citep{park2019} and \citep{tung2019}; however, these approaches have been proposed for the image domain. Video-based approaches~\citep{bhardwaj2019,zhang2018,garcia2018,liang2019} distill information between intermediate representations, e.g., video/frame activations or attention maps -- this is costly due to the high computational requirements of the teacher. By contrast, in our training scheme the teacher's similarities of the video pairs used during training can be pre-computed -- this allows training in large datasets in an unsupervised manner (i.e., without labels). Finally, these distillation methods end up with a single network that either offers compression or better performance -- by contrast, in the proposed framework, we are able to arrive at different accuracy/speed/storage trade-offs.
        
        \textbf{Network architectures:} We propose three student network architectures that are trained with Knowledge Distillation in an unsupervised manner on large unannotated datasets avoiding in this way overfitting (cf. Sect.~\ref{sec:distillation_vs_supervision}). Two fine-grained students are built based on our prior work in~\citep{kordopatis2019b}, with some essential adjustments to mitigate its limitations. A fine-grained attention student is developed using a more complex attention mechanism, which outperforms the Teacher when trained on the large unlabeled dataset. Also, a fine-grained binarization student is introduced with a binarization layer that has significantly lower storage requirements. Prior works have used binarization layers with coarse-grained approaches~\citep{liong2017,song2018,yuan2020}, but none learns a fine-grained similarity function based on binarized regional-level descriptors. Furthermore, a coarse-grained student is built. Its novelties are the use of a trainable region-level aggregation scheme -- unlike other works that extract frame-level descriptors -- and the combination of two aggregation components on frame-level that considers intra- and inter-video relations between frames. Prior works have employed a transformer encoder to capture intra-video frame relations \citep{shao2021}, or a NetVLAD to capture inter-video ones \citep{miech2017}; however, none combines the two components together.

%%%%%%%%%%%%%%%%%%%%%%%%%%%%%%%%%%%%%%%%%%%%%%%%%%%%%%%%%%%%%%%%%%%%%%%%%%%%%%%%
%%                                                                            %%
%%                           [Proposed Method]                                %%
%%                                                                            %%
%%%%%%%%%%%%%%%%%%%%%%%%%%%%%%%%%%%%%%%%%%%%%%%%%%%%%%%%%%%%%%%%%%%%%%%%%%%%%%%%
\section{Distill-and-Select}\label{sec:proposed_method}

    This section presents the Distill-and-Select (DnS) method for video retrieval. First, we describe the developed retrieval pipeline, which involves a fine-grained and a coarse-grained student network trained with Knowledge Distillation, and a selector network, acting as a re-ranking mechanism (Sect.~\ref{sec:retrieval}). Then, we discuss the network architectures/alternatives employed in our proposed approach that offer different performance-efficiency trade-offs (Sect.~\ref{sec:architectures}). Finally, the training processes followed for the training of the proposed networks are presented (Sect.~\ref{sec:training_process}).
    
    \begin{figure*}[t]
        \centering
        \includegraphics[width=\textwidth]{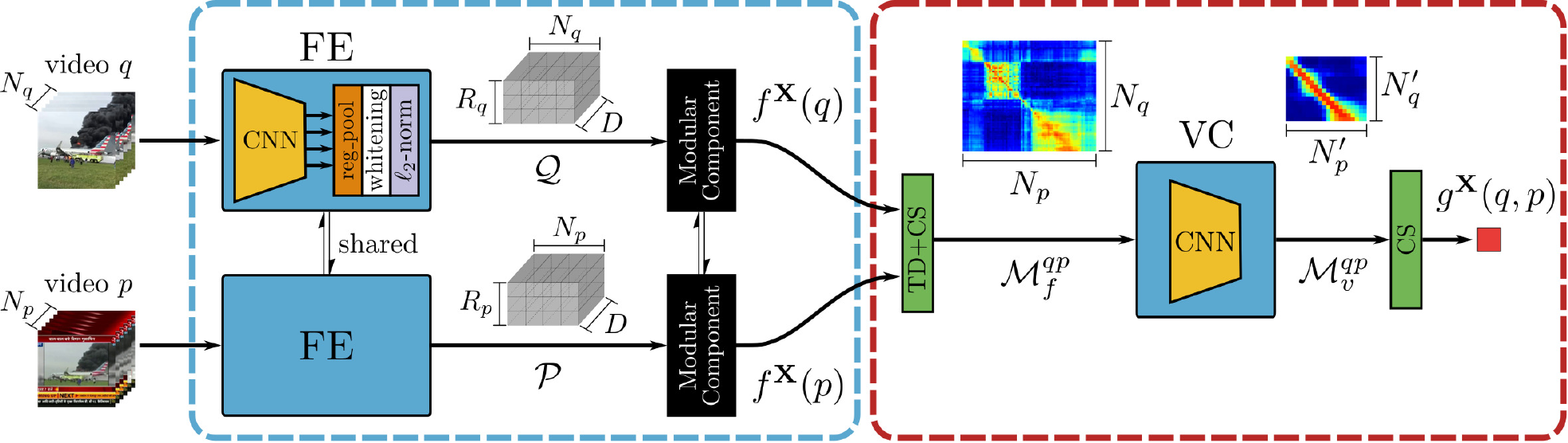}
        \caption{Illustration of ViSiL~\citep{kordopatis2019b} architecture used for the teacher and fine-grained students, i.e., $\textbf{X}$ can take three values: $\textbf{T}$, $\textbf{S}^f_\mathcal{A}$, and $\textbf{S}^f_\mathcal{B}$. During indexing, a 3D video tensor is extracted based on a Feature Extraction (FE) process, applying regional pooling, whitening, and $\ell^2$-normalization on the activations of a CNN. Then, a modular component is applied according to the employed network, i.e., an attention scheme for $\textbf{T}$ and $\textbf{S}^f_\mathcal{A}$, and a binarization layer for $\textbf{S}^f_\mathcal{B}$. During retrieval, the Tensor Dot (TD) followed by Chamfer Similarity (CS) are applied on the representations of a video pair to generate their frame-to-frame similarity matrix, which is propagated to a Video Comparator (VC) CNN that captures the temporal patterns. Finally, CS is applied again to derive a single video-to-video similarity score.}
        \label{fig:visil_arch}
    \end{figure*}
    
    \subsection{Approach overview}\label{sec:retrieval}
        Fig. \ref{fig:overview} depicts the DnS framework. It consists of three networks: (i) a coarse-grained student ($\textbf{S}^c$) that provides very fast retrieval speed but with low retrieval performance, (ii) a fine-grained student ($\textbf{S}^f$) that has high retrieval performance but with high computational cost, and (iii) a selector network ($\textbf{SN}$) that routes the similarity calculation of the video pairs and provides a balance between performance and time efficiency.
        
        Each video in the dataset is stored/indexed using three representations: (i) a spatio-temporal 3D tensor $f^{\textbf{S}^f}$ that is extracted (and then used at retrieval time) by the fine-grained student $\textbf{S}^f$, (ii) a 1D global vector $f^{\textbf{S}^c}$ that is extracted (and then used at retrieval time) by the coarse-grained student $\textbf{S}^c$, and (iii) a scalar $f^{\textbf{SN}}$ that summarises the similarity between different frames of the video in question that is extracted (and then used at retrieval time) by the selector network $\mathbf{SN}$. The indexing process that includes the feature extraction is illustrated within the blue box in Fig.~\ref{fig:overview}, \ref{fig:visil_arch}, \ref{fig:coarse_arch}, \ref{fig:selector} and is denoted as $f^{\textbf{X}}(\cdot)$ for each network $\textbf{X}$. At retrieval-time, given an input query-target video pair, the selector network sends to the coarse-grained student $\textbf{S}^c$ the global 1D vectors so that their similarity is rapidly estimated (i.e., as the dot product of the representations) $g^{\mathbf{S}^c}$. This coarse similarity and the self-similarity scalars for the videos in question are then given as input to the selector $\mathbf{SN}$, which takes a binary decision $g^{\mathbf{SN}}$ on whether the calculated coarse similarity needs to be refined by the fine-grained student. For the small percentage of videos that this is needed, the fine-grained network calculates the similarity $g^{\mathbf{S}^f}$ based on the spatio-temporal representations. The retrieval process that includes the similarity calculation is illustrated within the red box in Fig.~\ref{fig:overview}, \ref{fig:visil_arch}, \ref{fig:coarse_arch}, \ref{fig:selector} and is denoted as $g^{\textbf{X}}(\cdot, \cdot)$ for each network $\textbf{X}$.
        
        In practice, we apply the above process on every query-target video pair derived from a database, and a predefined percentage of videos with the largest confidence score calculated by the selector is sent to the fine-grained student for re-ranking. With this scheme, we achieve very fast retrieval with very competitive retrieval performance.

    \subsection{Network architectures}\label{sec:architectures}
        In this section, the architectures of all networks included in the DnS framework are discussed. First, the teacher network that is based on the ViSiL architecture is presented (\ref{sec:baseline_teacher}). Then, we discuss our student architectures, which we propose under a Knowledge Distillation framework that addresses the limitations introduced by the teacher; i.e., high resource requirements, both in terms of memory space for indexing, due to the region-level video tensors, and computational time for retrieval, due to the fine-grained similarity calculation. More precisely, three students are proposed, two fine-grained and one coarse-grained variant, each providing different benefits. The fine-grained students are both using the ViSiL architecture. The first fine-grained student simply introduces more trainable parameters, leading to better performance with similar computational and storage requirements to the teacher (\ref{sec:attention_student}). The second fine-grained student optimizes a binarization function that hashes features into a Hamming space and has very low storage space requirements for indexing with little performance sacrifice (\ref{sec:binarization_student}). The third coarse-grained student learns to aggregate the region-level feature vectors in order to generate a global video-level representation and needs considerably fewer resources for indexing and retrieval but at notable performance loss (\ref{sec:binarization_student}). Finally, we present the architecture of the selector network for indexing and retrieval (\ref{sec:binarization_student}). Our framework operates with a specific combination of a fine-grained and coarse-grained student and a selector network. Each combination achieves different trade-offs between retrieval performance, storage space, and computational time.
        
        \subsubsection{Baseline Teacher ($\textbf{T}$)}\label{sec:baseline_teacher}
        
            Here, we will briefly present the video similarity learning architecture that we employ as the teacher and which builds upon the ViSiL~\citep{kordopatis2019b} architecture (Fig.~\ref{fig:visil_arch}). 
            
            \bigskip
            \noindent\textbf{Feature extraction/Indexing} ($f^{\textbf{T}}$): Given an input video, we first extract region-level features from the intermediate convolution layers \citep{kordopatis2017a} of a backbone CNN architecture by applying region pooling \citep{tolias2016} on the feature maps. These are further PCA-whitened~\citep{jegou2012} and $\ell^2$-normalized. We denote the aforementioned process as Feature Extraction (FE), and we employ it in all of our networks. FE is followed by a modular component, as shown in Fig.~\ref{fig:visil_arch}, that differs for each fine-grained student. In the case of the teacher, an attention mechanism is employed imposing that frame regions are weighted based on their saliency via a visual attention mechanism over region vectors based on an $\ell^2$-normalized context vector. The context vector is a trainable vector $\textbf{u}\in\mathbb{R}^{D}$ that weights each region vector independently based on their dot-product. It is learned through the training process. Also, no fully-connected layer is employed to transform the region vectors for the attention calculation. We refer to this attention scheme as $\ell^2$-attention. The output representation of an input video $x$ is a region-level video tensor $\mathcal{X}\in\mathbb{R}^{N_x\times R_x\times D}$, where $N_x$ is the number of frames, $R_x$ is the number of regions per frame, and $D$ is the dimensionality of the region vectors -- this is the output of the indexing process, and we denote it as $f^{\textbf{T}}(x)$.
            
            \bigskip
            \noindent\textbf{Similarity calculation/Retrieval} ($g^{\textbf{T}}$): At retrieval time, given two videos, $q$ and $p$, with $N_q$ and $N_p$ number of frames and $R_q$ and $R_p$ regions per frame, respectively, for every pair of frames, we first calculate the frame-to-frame similarity based on the similarity of their region vectors. More precisely, to calculate the \emph{frame-to-frame} similarity on videos $q$ and $p$, we calculate the Tensor Dot combined with Chamfer Similarity on the corresponding video tensors $f^{\textbf{T}}(q)=\mathcal{Q}\in\mathbb{R}^{N_q\times R_q\times D}$ and $f^{\textbf{T}}(p)=\mathcal{P}\in\mathbb{R}^{N_p\times R_p\times D}$ as follows
            \begin{equation}\label{eq:similarity_matrix}
                \mathcal{M}_f^{qp} = \frac{1}{R_q}\sum_{i=1}^{R_q} \max_{1\leq j\leq R_p} \mathcal{Q} \bigcdot _{(3,1)} \mathcal{P}^\top(\cdot,i,j,\cdot),
            \end{equation}
            where $\mathcal{M}_f^{qp}\in\mathbb{R}^{N_q\times N_p}$ is the output frame-to-frame similarity matrix, and the Tensor Dot axes indicate the channel dimension of the corresponding video tensors. Also, the Chamfer Similarity is implemented as a max-pooling operation followed by an average-pooling on the corresponding dimensions. This process leverages the geometric information captured by region vectors and provides some degree of spatial invariance. Also, it is worth noting that this frame-to-frame similarity calculation process is independent of the number of frames and region vectors; thus, it can be applied on any video pair with arbitrary sizes and lengths. 
            
            To calculate the \emph{video-to-video} similarity, the generated similarity matrix $\mathcal{M}_f^{qp}$ is fed to a Video Comparator (VC) CNN module (Fig.~\ref{fig:visil_arch}), which is capable of learning robust patterns of within-video similarities. The output of the network is the refined similarity matrix $\mathcal{M}_v^{qp}\in\mathbb{R}^{N'_q\times N'_p}$. In order to calculate the final video-level similarity for two input videos $q,p$, i.e., $g^{\textbf{T}}(q,p)$, the hard tanh ($\operatorname{Htanh}$) activation function is applied on the values of the aforementioned network output followed by Chamfer Similarity in order to obtain a single value, as follows
            \begin{equation}\label{eq:video_level_similarity}
                g^{\textbf{T}}(q,p) = \frac{1}{N'_q}\sum_{i=1}^{N'_q}\max_{1\leq j\leq N'_p} 
                \operatorname{Htanh}\left(  \mathcal{M}_v^{qp}(i,j)\right).
            \end{equation}
            In that way, the VC takes temporal consistency into consideration by applying learnable convolutional operations on the frame-to-frame similarity matrix. Those enforce local temporal constraints while the Chamfer-based similarity provides invariance to global temporal transformations. Hence, similarly to the frame-to-frame similarity calculation, this process is a trade-off between respecting the video-level structure and being invariant to some temporal differences.

        \subsubsection{Fine-grained attention student ($\textbf{S}^f_\mathcal{A}$)}\label{sec:attention_student}
            The first fine-grained student adopts the same architecture as the teacher (Sect.~\ref{sec:baseline_teacher}, Fig.~\ref{fig:visil_arch}), but uses a more complex attention scheme in the modular component, employed for feature weighting, as proposed in~\citep{yang2016}.
            
            \bigskip
            \noindent\textbf{Feature extraction/Indexing} ($f^{\textbf{S}^f_\mathcal{A}}$): The Feature Extraction (FE) process is used to extract features, similar to the teacher. In the modular component shown in Fig.~\ref{fig:visil_arch}, we apply an attention weighting scheme as follows. Given a region vector $\textbf{r}\colon\mathcal{X}(i,j,\cdot)\in\mathbb{R}^D$, where $i=1,\ldots,N_x$, $j=1,\ldots,R_x$, a non-linear transformation is applied, which is implemented as a fully-connected layer with tanh activation function, to form a hidden representation $\textbf{h}$. Then, the attention weight is calculated as the dot product between $\textbf{h}$ and the context vector $\textbf{u}$, followed by the sigmoid function, as
            \begin{equation}\label{eq:attention_mechanism_complex}
                \begin{aligned}
                    & \textbf{h} = \text{tanh}(\textbf{r}\cdot W_a + b_a),  \\
                    & \alpha = \text{sig}(\textbf{u}\cdot\textbf{h}), \\
                    & \textbf{r}' = \alpha\textbf{r},
                \end{aligned}
            \end{equation}
            where $W_a\in\mathbb{R}^{D\times D}$ and $b_a\in \mathbb{R}^{D}$ are the weight and bias parameters of the hidden layer of the attention module, respectively, and $\text{sig}(\cdot)$ denotes the element-wise sigmoid function. We will be referring to this attention scheme as $h$-attention. The resulting 3D representation is the indexing output $f^{\textbf{S}^f_\mathcal{A}}(x)$ for an input video $x$.
            
            \bigskip
            \noindent\textbf{Similarity calculation/Retrieval} ($g^{\textbf{S}^f_\mathcal{A}}$): To calculate similarity between two videos, we build the same process as for the teacher, i.e., we employ a Video Comparator (VC) and use the same frame-to-frame and video-to-video functions to derive $g^{\textbf{S}^f_\mathcal{A}}(q,p)$ for two input videos $q,p$ (Fig.~\ref{fig:visil_arch}).
            
            In comparison to the teacher, this student a) has very similar storage requirements, since in both cases, the videos are stored as non-binary spatio-temporal features, b) has similar computational cost, since the additional attention layer introduces only negligible overhead, and c) typically reaches better performance, since it has slightly higher capacity and can be trained in a much larger, unlabelled dataset.

        \subsubsection{Fine-grained binarization student ($\textbf{S}^f_\mathcal{B}$)}\label{sec:binarization_student}
            The second fine-grained student also adopts the same architecture as the teacher (Sect.~\ref{sec:baseline_teacher}, Fig.~\ref{fig:visil_arch}), except for the modular component where a binarization layer is introduced, as discussed below.
            
            \bigskip
            \noindent\textbf{Feature extraction/Indexing} ($f^{\textbf{S}^f_\mathcal{B}}$): $f^{\textbf{S}^f_\mathcal{B}}$ is the part of the indexing of the student $\textbf{S}^f_\mathcal{B}$ that extracts a binary representation for an input video that will be stored and used at retrieval time. It uses the architecture of the teacher, where the modular component is implemented as a binarization layer (Fig.~\ref{fig:visil_arch}). This applies a binarization function $b\colon\mathbb{R}^D\to\{-1, 1\}^L$ that hashes the region vectors $\textbf{r}\in\mathbb{R}^D$ to binary hash codes $\textbf{r}_\mathcal{B}\in\{-1,1\}^L$ as
            \begin{equation}\label{eq:binarization_function}
                b(\textbf{r}) = \text{sgn}\left(\textbf{r}\cdot W_\mathcal{B}\right),
            \end{equation}
            where $W_\mathcal{B}\in \mathbb{R}^{D\times L}$ denote the learnable weights and $\text{sgn}(\cdot)$ denotes the element-wise sign function.
            
            However, since $\text{sgn}$ is not a differentiable function, learning binarization parameters via backpropagation is not possible. To address this, we propose an approximation of the sign function under the assumption of small uncertainty in its input. More specifically, let $\text{sgn}\colon x\mapsto\{\pm1\}$, where $x$ is drawn from a uni-variate Gaussian distribution with given mean $\mu$ and fixed variance $\sigma^2$, i.e., $x\sim\mathcal{N}(\mu,\sigma^2)$. Then, the expected value\footnote{$\EX[\text{sgn}(x)]=1\cdot\mathbb{P}(x>0)+(-1)\cdot\mathbb{P}(x<0)=1-2\cdot\mathbb{P}(x<0)=1-2\mathbb{P}\left(z<-\frac{\mu}{\sqrt{2\sigma^2}}\right)\implies\EX[\text{sgn}(x)]=1-2\Phi\left(\frac{-\mu}{\sqrt{2\sigma^2}}\right)=\operatorname{erf}\left(\frac{\mu}{\sqrt{2\sigma^2}}\right)$, where $z\sim\mathcal{N}(0,1)$ denotes the standard Gaussian and $\Phi$ its CDF.} of the sign of $x$ is given analytically as follows
            \begin{equation}\label{eq:bin_activation_function}
                \EX[\text{sgn}(x)] = \operatorname{erf}\left(\frac{\mu}{\sqrt{2\sigma^2}}\right),
            \end{equation}
            where $\text{erf}(\cdot)$ denotes the error function. This is differentiable and therefore can serve as an activation function on the binarization parameters, that is,
            \begin{equation}\label{eq:binarization_function_train}
                b(\textbf{r}) = \operatorname{erf}\left(\frac{\textbf{r}\cdot W_\mathcal{B}}{\sqrt{2\sigma^2}}\right),
            \end{equation}
            where we use as variance an appropriate constant value (empirically set to $\sigma=10^{-3}$). During training, we use (\ref{eq:binarization_function_train}), while during evaluation and hash code storage we use (\ref{eq:binarization_function}). After applying this operation to an arbitrary video $x$ with $N_x$ frames and $R_x$ regions, we arrive at a binary tensor $\mathcal{X}_\mathcal{B}\in\{\pm1\}^{N_x\times R_x\times L}$, which is the indexing output $f^{\textbf{S}^f_\mathcal{B}}(x)=\mathcal{X}_\mathcal{B}$ used by this student.
            
            \bigskip
            \noindent\textbf{Similarity calculation/Retrieval} ($g^{\textbf{S}^f_\mathcal{B}}$): In order to adapt the similarity calculation processes with the binarization operation, the Hamming Similarity (HS) combined with Chamfer Similarity is employed as follows. Given two videos $q,p$ with $f^{\textbf{S}^f_\mathcal{B}}(q)=\mathcal{Q}_\mathcal{B}\in\{\pm1\}^{N_q\times R_q\times L}$ and $f^{\textbf{S}^f_\mathcal{B}}(p)=\mathcal{P}_\mathcal{B}\in\{\pm1\}^{N_p\times R_p\times L}$ their binary tensors, respectively, we first calculate the HS between the two tensors with the use of Tensor Dot to calculate the similarity of all region pair combinations of the two videos and then apply Chamfer Similarity to derive the frame-to-frame similarity matrix $\mathcal{M}_\mathcal{B}^{qp}\in\mathbb{R}^{N_q\times N_p}$. That is,
            \begin{equation}\label{eq:hamming_similarity_matrix}
                \begin{aligned}
                    & HS^{qp} = \mathcal{Q}_\mathcal{B} \bigcdot _{(3,1)} \mathcal{P}_\mathcal{B}^\top / L, \\
                    & \mathcal{M}_\mathcal{B}^{qp} = \frac{1}{R_q}\sum_{i=1}^{R_q} \max_{1\leq j\leq R_p} HS^{qp} (\cdot, i, j, \cdot),
                \end{aligned}
            \end{equation}
            Finally, a Video Comparator (VC) is applied on the frame-to-frame similarity matrices in order to calculate the final video-to-video similarity, similarly to (\ref{eq:video_level_similarity}) in the original teacher (Fig.~\ref{fig:visil_arch}) -- this is denoted as $g^{\textbf{S}^f_\mathcal{B}}(q,p)$ for two input videos $q,p$.
            
            In comparison to the teacher, this student a) has remarkably lower storage requirements, since the binary spatio-temporal representations are 32 times smaller than the corresponding float ones (full precision), b) has similar computational cost, as the architecture is very similar, and c) reaches better performance since it is trained at a larger (despite being unlabelled) dataset. Note that this student only uses a binary input but is not a binarized network.

        \subsubsection{Coarse-grained student ($\textbf{S}^c$)}\label{sec:coarse_student}
\begin{figure}[t]
                \centering
                \includegraphics[width=\linewidth]{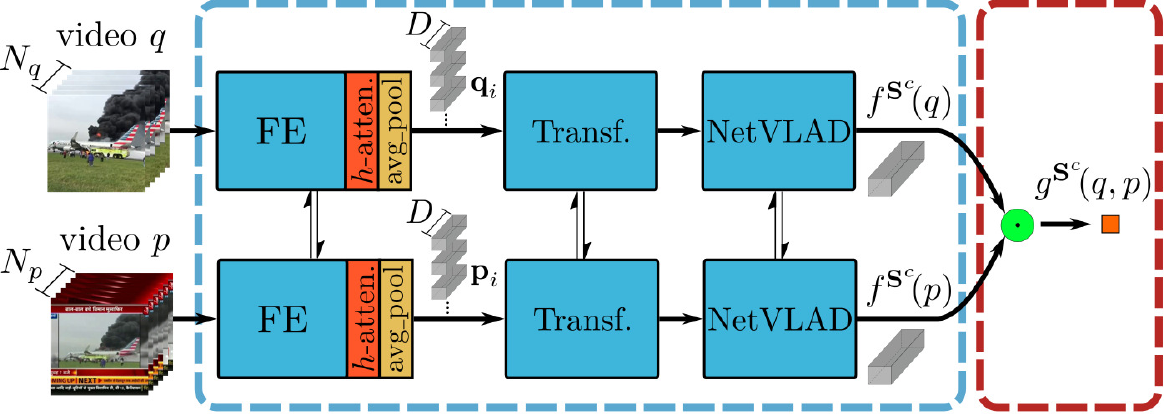}
                \caption{Illustration of the architecture of the coarse-grained student $\textbf{S}^c$, consisting of three main components. During indexing, the FE process with attention weighting and average pooling is applied to extract frame-level features. Then, they are processed by a Transformer network and aggregated to 1D vectors by a NetVLAD module. During retrieval, the video similarity derives from a simple dot product between the extracted representations.}
                \label{fig:coarse_arch}
            \end{figure}

            The coarse-grained student introduces an architecture that extracts video-level representations that are stored and can be subsequently used at retrieval time so as to rapidly estimate the similarity between two videos as the cosine similarity of their representations. An overview of the coarse student is shown in Fig.~\ref{fig:coarse_arch}.
            
            \bigskip
            \noindent\textbf{Feature extraction/Indexing} ($f^{\textbf{S}^c}$): The proposed coarse-grained student comprises of three components. First, we extract weighted region-level features with Feature Extraction (FE), using the attention module given by (\ref{eq:attention_mechanism_complex}), and then average pooling is applied across the spatial dimensions of the video tensors lead to frame-level representations for the videos; i.e., $\textbf{x}_i=\frac{1}{R_x} \sum_{k=1}^{R_x} \alpha_k\textbf{r}_k$, where $\textbf{x}_i\in\mathbb{R}^D$ is the frame-level vector of the $i$-th video frame, $R_x$ is the number of regions, and $\alpha_k$ is the attention weight calculated by (\ref{eq:attention_mechanism_complex}). In that way, we apply a trainable scheme to aggregate the region-level features that focuses on the information-rich regions. Second, a transformer~\citep{vaswani2017} network architecture is used to derive frame-level representations that capture long-term dependencies within the frame sequence, i.e., it captures the intra-video relations between frames. Following~\cite{shao2021}, the encoder part of the Transformer architecture is used, which is composed of a multi-head self-attention mechanism and a feed-forward network. Finally, a NetVLAD~\citep{arandjelovic2016} module aggregates the entire video to a single vector representation~\citep{miech2017}. This component learns a number of cluster centers and a soft assignment function through the training process, considering all videos in the training dataset. Therefore, it can be viewed as it encodes the inter-video relations between frames. Given input a video $x$, the output $f^{\textbf{S}^c}(x)$ is a 1D video-level vector that is indexed and used by the coarse-grained student during retrieval.
            
            \bigskip
            \noindent\textbf{Similarity calculation/Retrieval} ($g^{\textbf{S}^c}$): Once feature representations have been extracted, the similarity calculation is a simple dot product between the 1D vectors of the compared videos, i.e., $g^{\textbf{S}^c}(q,p)=f^{\mathbf{S}^c}(q)\cdot f^{\mathbf{S}^c}(p)$ for two input videos $q,p$.
            
            In comparison to the original teacher, this student a) has remarkably lower storage requirements for indexing, since it stores video-level representations instead of spatio-temporal ones, b) has significantly lower computational cost at retrieval time, since the similarity is calculated with a single dot-product between video-level representations, and c) has considerably lower performance, since it does not model spatio-temporal relations between videos during similarity calculation.

            \begin{figure*}[t]
                \centering
                \includegraphics[width=\textwidth]{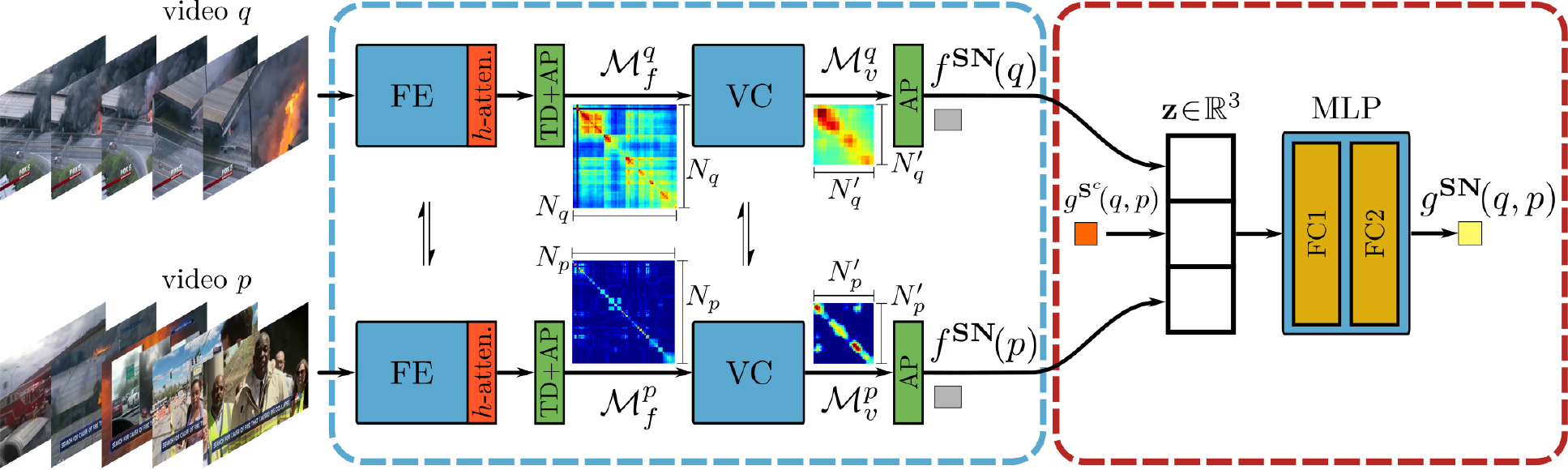}
                \caption{Illustration of the Selector Network architecture. During indexing, the self-similarity of the videos is calculated according to the following scheme. First, region-level attention-weighted features are extracted. Then, the frame-to-frame self-similarity matrix derives with a Tensor Dot (DT) and Average Pooling (AP), which is propagated to a VC module to capture temporal patterns. The final self-similarity is calculated based on an AP on the VC output. During retrieval, given a video pair, a 3-dimensional vector is composed by the self-similarity of each video and their similarity calculated by the $\textbf{S}^c$. The feature vector is fed to an MLP to derive a confidence score.}
                \label{fig:selector}
            \end{figure*}
        
        \subsubsection{Selector network ($\textbf{SN}$)}\label{sec:selector_network}
            
            In the proposed framework, at retrieval time, given a pair of videos, the role of the selector is to decide whether the similarity that is calculated rapidly based on the stored coarse video-level representations is accurate enough (i.e., similar to what a fine-grained student would give), or whether a fine-grained similarity, based on the spatio-temporal, fine-grained representations needs to be used, and a new, refined similarity measure needs to be estimated. Clearly, this decision needs to be taken rapidly and with a very small additional storage requirement for each video.
            
            The proposed selector network is shown in Fig.~\ref{fig:selector}. At retrieval time, a simple Multi-Layer Perceptron (MLP) takes as input a three dimensional vector, $\mathbf{z}\in\mathbb{R}^3$, with the following features: a) the similarity between a pair of videos $q,p$, as calculated by $\textbf{S}^c$ (Sect.~\ref{sec:coarse_student}), and b) the fine-grained self-similarities $f^{\textbf{SN}}(q)$ and $f^{\textbf{SN}}(p)$, calculated by a trainable NN (Fig.~\ref{fig:selector}). Since $f^{\textbf{SN}}(x)$ depends only on video $x$, it can be stored together with the representations of the video $x$ with negligible storage cost. Having $f^{\textbf{SN}}(q)$ and $f^{\textbf{SN}}(p)$ pre-computed, and $g^{\textbf{S}^c}$ rapidly computed by the coarse-grained student, the use of selector at retrieval time comes at a negligible storage and computational cost. Both the self-similarity function $f^{\textbf{SN}}$ that extracts features at indexing time and the MLP that takes the decision at retrieval time, which are parts of the Selector Network $\textbf{SN}$, are jointly trained.
            
            In what follows, we describe the architecture of the selector, starting from the network that calculates the fine-grained similarity $f^{\textbf{SN}}$. This is a modified version of the ViSiL architecture that aims to derive a measure that captures whether there is large spatio-temporal variability in its content. This is expected to be informative on whether the fine-grained student needs to be invoked. The intuition is that for videos with high $f^{\textbf{SN}}$, i.e., not high spatio-temporal variability, their video-level representations are sufficient to calculate their similarity, i.e., the similarity estimated by the coarse-grained student is accurate enough. 
            
            \bigskip
            \noindent\textbf{Feature extraction/Indexing} ($f^{\textbf{SN}}$): Given a video $x$ as input, features are extracted based on the Feature Extraction (FE), using the attention module as in (\ref{eq:attention_mechanism_complex}), to derive a video tensor $\mathcal{X}\in\mathbb{R}^{N_x\times R_x\times D}$. Then, the frame-to-frame self-similarity matrix is calculated, as
            \begin{equation}\label{eq:self_similarity_matrix}
                \mathcal{M}_f^{x} = \frac{1}{R_x^2}\sum_{i=1}^{R_x} \sum_{j=1}^{R_x} \mathcal{X} \bigcdot _{(3,1)} \mathcal{X}^\top(\cdot,i,j,\cdot),
            \end{equation}
            where, $\mathcal{M}_f^{x}\in\mathbb{R}^{N_x\times N_x}$ is the symmetric frame-to-frame self-similarity matrix. Note that (\ref{eq:self_similarity_matrix}) is a modified version of (\ref{eq:similarity_matrix}), where the Chamfer Similarity is replaced by the average operator. In this case, we calculate the average similarity of a region with all other regions in the same frame -- the use of Chamfer Similarity would have resulted in estimating the similarity of a region with the most similar region in the current frame, that is itself.
            
            Similarly, a Video Comparator (VC) CNN network is employed (same as ViSiL, Fig.~\ref{fig:visil_arch}) that is fed with the self-similarity matrix in order to extract the temporal patterns and generate a refined self-similarity matrix $\mathcal{M}_v^{x}\in\mathbb{R}^{N'_x\times N'_x}$. To extract a final score (indexing output) that captures self-similarity, we modify (\ref{eq:video_level_similarity}) as
            \begin{equation}\label{eq:self_similarity}
                f^{\textbf{SN}}(x) = \frac{1}{N'_x{}^2}\sum_{i=1}^{N'_x}\sum_{j=1}^{N'_x}
                \mathcal{M}_v^{x}(i,j),
            \end{equation}
            that is, the average of the pair-wise similarities of all video frames. Note that we also do not use the hard tanh activation function, as we empirically found that it is not needed. 
            
            \bigskip
            \noindent\textbf{Confidence calculation/Retrieval} ($g^{\textbf{SN}}$): Given a pair of videos and their similarity predicted by the $\textbf{S}^c$, we retrieve the indexed self-similarity scores, and then we concatenate them with the $\textbf{S}^c$ similarity, forming a three-dimensional vector $\textbf{z}\in\mathbb{R}^3$ for the video pair, as shown in Fig.~\ref{fig:selector}. This vector is given as input to a two-layer MLP using Batch Normalization~\citep{ioffe2015} and ReLU~\citep{krizhevsky2012} activation functions. For an input video pair $q,p$, the retrieval output $g^{\textbf{SN}}(q, p)$ is the confidence score of the selector network that the fine-grained student needs to be invoked.

    \subsection{Training process}\label{sec:training_process}
        
        \begin{figure}[t]
            \centering
            \includegraphics[width=\linewidth]{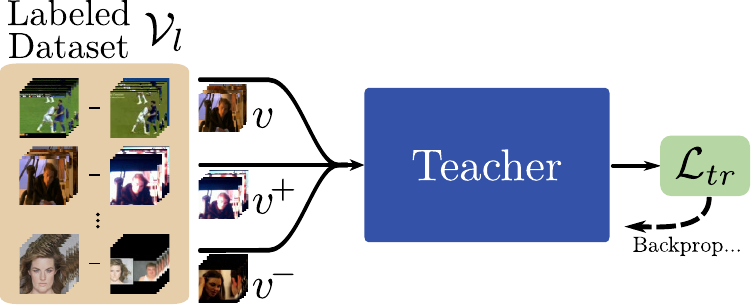}
            \caption{Illustration of the training process of the teacher networks. It is trained with supervision with video triplets derived from a labelled dataset, minimizing triplet loss.}
            \label{fig:teacher_training}
        \end{figure}
    
        In this section, we go through the details of the procedure followed for the training of the underlying networks of the proposed framework, i.e., the teacher, the students, and the selector.
        
        \subsubsection{Teacher training}\label{sec:teacher_training}
            The teacher network is trained with supervision on a labelled video dataset $\mathcal{V}_l$, as shown in Fig.~\ref{fig:teacher_training}. The videos are organized in triplets $(v,v^+,v^-)$ of an anchor, a positive (relevant), and a negative (irrelevant) video, respectively, where $v,v^+,v^-\in\mathcal{V}_l$, and the network is trained with the \textit{triplet loss}
            \begin{equation}\label{eq:triplet_loss}
                \mathcal{L}_{tr} = \max\left(0, g^\textbf{T}(v, v^-) - g^\textbf{T}(v, v^+) + \gamma\right),
            \end{equation}
            where $\gamma$ is a margin hyperparameter. In addition, a similarity regularization function is used that penalizes high values in the input of hard tanh that would lead to saturated outputs. Following other works, we use data augmentation (i.e., color, geometric, and temporal augmentations) on the positive samples $v^+$.
        
        \subsubsection{Student training}\label{sec:student_training}
\begin{figure}[t]
                \centering
                \includegraphics[width=0.95\linewidth]{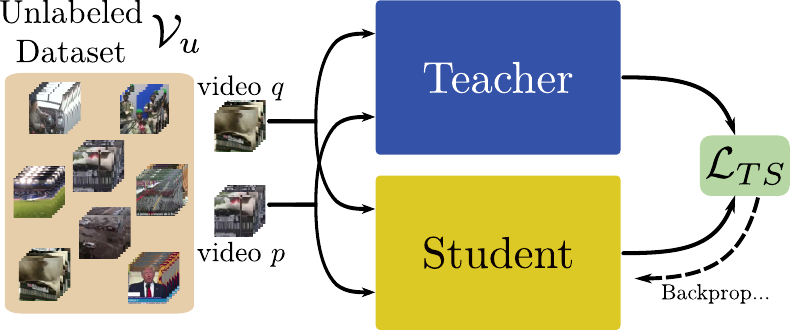}
                \caption{Illustration of the training process of the student networks. They are trained on an unlabelled dataset by minimizing the difference between their video similarity estimations and the ones calculated by the teacher network.}
                \label{fig:student_training}
            \end{figure}

            An overview of the student training process is illustrated in Fig.~\ref{fig:student_training}. Let $\mathcal{V}_u=\{v_1,v_2,\ldots,v_n\}$ be a collection of unlabelled videos and $g^\textbf{T}(q,p)$, $g^\textbf{S}(q, p)$ be the similarities between videos $q,p\in\mathcal{V}_u$, estimated by a teacher network $\textbf{T}$ and a student network $\textbf{S}$, respectively. $\textbf{S}$ is trained so that $g^\textbf{S}$ approximates $g^\textbf{T}$, with the $L_1$ loss\footnote{We have experimented with other losses, i.e., $L_2$ and Huber loss, with no considerable performance difference.}, that is,
            \begin{equation}
                \mathcal{L}_{TS} = \left\lVert g^\textbf{T}(q, p) - g^\textbf{S}(q, p)\right\rVert_1.
            \end{equation}
            
            Note that the loss is defined on the output of the teacher. This allows for a training process in which the scores of the teacher are calculated for a number of pairs in the unlabelled dataset only once, and then being used as targets for the students. This is in contrast to methods where the loss is calculated on intermediate features of $\textbf{T}$ and $\textbf{S}$, and cannot, thus, scale to large-scale datasets as they have considerable storage and/or computational/memory requirements. In this setting, the selection of the training pairs is crucial. Since it is very time consuming to apply the teacher network $\textbf{T}$ to every pair of videos in the dataset ($O(n^2)$ complexity) and randomly selecting videos would result in mostly pairs with low similarity scores, here, we follow~\cite{kordopatis2019a} and generate a graph to extract its connected components, which are considered as video clusters. Each video included in a video cluster is considered as an anchor, and we form pairs with the videos belonging to the same cluster, which are treated as positive pairs. Also, based on the anchor video, we form pairs with the 50 most similar videos that belong to the other clusters and the 50 most similar videos that belong to no cluster, which are treated as negative pairs. At each epoch, one positive and one negative pair are selected for each anchor video to balance their ratio.

        \subsubsection{Selector training}\label{sec:selector_training}
            
            Typically, the similarity between two videos $q$ and $p$ that is estimated by a fine-grained student $\textbf{S}^f$, leads to better retrieval scores than the one estimated by the coarse-grained student $\textbf{S}^c$. However, for some video pairs, the difference between them (i.e., $\lVert g^{\textbf{S}^c}(q,p) - g^{\textbf{S}^f}(q,p)\rVert_1$) is small and, therefore, having negligible effect to the ranking and on whether the video will be retrieved or not. The selector is a network that is trained to distinguish between those video pairs, and pairs of videos that exhibit large similarity differences. For the former, only the coarse-grained student $\textbf{S}^c$ will be used; for the latter, the fine-grained student $\textbf{S}^f$ will be invoked.
            
\begin{figure}[t]
                \centering
                \includegraphics[width=\linewidth]{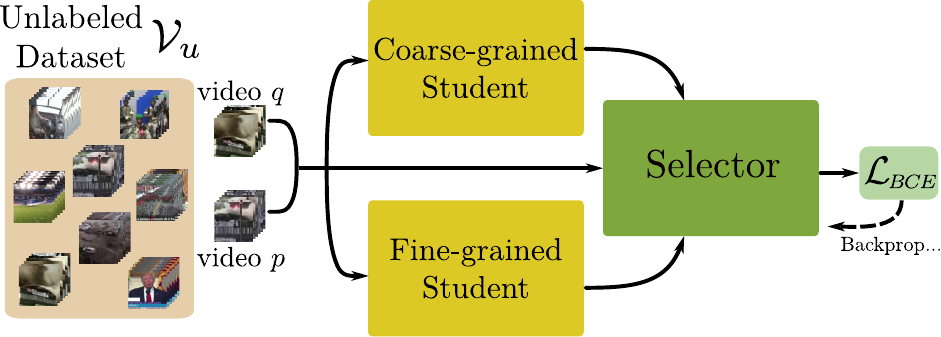}
                \caption{Illustration of the training process of the selector network. It is trained on an unlabelled dataset, exploiting the similarities calculated by a coarse- and fine-grained student. Note that the fine-grained student is applied on all video pairs only during training time. During retrieval, only a portion of the dataset is sent to it.}
                \label{fig:selector_training}
            \end{figure}

            The selector network is trained as a binary classifier, with binary labels obtained by setting a threshold $t$ on $\lVert g^{\textbf{S}^c}(q,p) - g^{\textbf{S}^f}(q,p)\rVert_1$, that is,
            \begin{equation}\label{eq:label}
                l(q,p) =
                \begin{cases}
                    1  &  \text{if}\ \left\lVert g^{\textbf{S}^c}(q,p) - g^{\textbf{S}^f}(q,p)\right\rVert_1 > t, \\
                    0  &  \text{otherwise}.
                \end{cases}
            \end{equation}
            Video pairs are derived from $\mathcal{V}_u$, and Binary Cross-Entropy is used as a loss function, as shown in Fig.~\ref{fig:selector_training}. We use the same mining process used for the student training, and at each epoch, a fixed number of video pairs is sampled for the two classes. We reiterate here that the selector is trained in an end-to-end manner, i.e., both the self-similarity feature extraction network $f^{\textbf{SN}}$, given by (\ref{eq:self_similarity}), and the decision-making MLP (Fig.~\ref{fig:selector}) are optimized jointly during training.
            
%%%%%%%%%%%%%%%%%%%%%%%%%%%%%%%%%%%%%%%%%%%%%%%%%%%%%%%%%%%%%%%%%%%%%%%%%%%%%%%%
%%                                                                            %%
%%                           [Evaluation Setup]                               %%
%%                                                                            %%
%%%%%%%%%%%%%%%%%%%%%%%%%%%%%%%%%%%%%%%%%%%%%%%%%%%%%%%%%%%%%%%%%%%%%%%%%%%%%%%%
\section{Evaluation Setup}\label{sec:evaluation_setup}
    In this section, we present the datasets (Sect.~\ref{sec:datasets}), evaluation metrics (Sect.~\ref{sec:metric}), and implementation details (Sect.~\ref{sec:implementation_details}) adopted during the experimental evaluation of the proposed framework. 

    % +-------------------------------------------------------------------------+ %
    % |                                                                         | %
    % |                             Datasets                                    | %
    % |                                                                         | %
    % +-------------------------------------------------------------------------+ %
    \subsection{Datasets}\label{sec:datasets}
        
        \subsubsection{Training datasets}

            \textbf{VCDB} \citep{jiang2014} was used as the training dataset to generate triplets for the training of the teacher model. The dataset consists of videos derived from popular video platforms (i.e., YouTube and Metacafe) and has been developed and annotated as a benchmark for partial copy detection. It contains two subsets, namely, the core and the distractor subsets. The former one contains 28 discrete sets composed of 528 videos with over 9,000 pairs of copied segments. The latter subset is a corpus of approximately 100,000 randomly collected videos that serve as distractors.
            
            \textbf{DnS-100K} is the dataset collected for the training of the students. We followed the collection process from our prior work~\citep{kordopatis2019a} for the formation of the FIVR-200K dataset in order to collect a large corpus of videos with various relations between them. First, we built a collection of the major news events that occurred in recent years by crawling Wikipedia's ``Current Event'' page\footnote{\url{https://en.wikipedia.org/wiki/Portal:Current_events}}. To avoid overlap with FIVR-200K, where the crawling period was from 2013-2017, we only considered the news events from the years 2018-2019. Then, we retained only the news events associated with armed conflicts and natural disasters by filtering them based on their topic. Afterwards, the public YouTube API\footnote{\url{https://developers.google.com/youtube/}} was used to collect videos by providing the event headlines as queries. The results were filtered to contain only videos published at the corresponding event start date and up to one week after the event. At the end of this process, we had collected a corpus of 115,792 videos. Following the mining scheme described in Sect.~\ref{sec:student_training}, we arrived at 21,997 anchor videos with approximately 2.5M pairs.

        \subsubsection{Evaluation datasets}
    
            \textbf{FIVR-200K} \citep{kordopatis2019a} was used as a benchmark for Fine-grained Incident Video Retrieval (FIVR). It consists of 225,960 videos collected based on the 4,687 events and 100 video queries. It contains video-level annotation labels for near-duplicate (ND), duplicate scene (DS), complementary scene (CS), and incident scene (IS) videos. FIVR-200K includes three different tasks: a) the Duplicate Scene Video Retrieval (DSVR) task, where only videos annotated with ND and DS are considered relevant, b) the Complementary Scene Video Retrieval (CSVR) task, which accepts only the videos annotated with ND, DS, or CS as relevant, and c) Incident Scene Video Retrieval (ISVR) task, where all labels are considered relevant. For quick comparisons of different configurations, we also used \textbf{FIVR-5K}, a subset of FIVR-200K, as provided in~\citep{kordopatis2019b}.
            
            \textbf{CC\_WEB\_VIDEO} \citep{wu2007} simulates the Near-Duplicate Video Retrieval (NDVR) problem. It consists of 24 query sets and 13,112 videos. The collection consists of a sample of videos retrieved by submitting 24 popular text queries to popular video sharing websites, i.e., YouTube, Google Video, and Yahoo! Video. For every query, a set of video clips was collected, and the most popular video was considered to be the query video. Subsequently, all videos in the video set were manually annotated based on their near-duplicate relation to the query video. We also use the `cleaned' version, as provided in~\citep{kordopatis2019b}.
            
            \textbf{SVD} \citep{jiang2019} was used for the NDVR problem, tailored for short videos in particular. It consists of 562,013 short videos crawled from a large video-sharing website, namely, Douyin\footnote{\url{http://www.douyin.com}}. The average length of the collected videos is 17.33 seconds. The videos with more than 30,000 likes were selected to serve as queries. Candidate videos were selected and annotated based on a three-step retrieval process. A large number of probably negative unlabelled videos were also included to serve as distractors. Hence, the final dataset consists of 1,206 queries with 34,020 labelled video pairs and 526,787 unlabelled videos. The queries were split into two sets, i.e., training and test set, with 1,000 and 206 queries, respectively. In this paper, we only use the test set for the evaluation of the retrieval systems.
            
            \textbf{EVVE} \citep{revaud2013} was designed for the Event Video Retrieval (EVR) problem. It consists of 2,375 videos and 620 queries. The main task on this dataset is the retrieval of all videos that capture the event depicted by a query video. The dataset contains 13 major events that were provided as queries to YouTube. Each event was annotated by one annotator, who first produced a precise definition of the event. However, we managed to download and process only 1906 videos and 504 queries (that is, $\approx$80\% of the initial dataset) due to the unavailability of the remaining ones.

    % +-------------------------------------------------------------------------+ %
    % |                                                                         | %
    % |                          Evaluation metric                              | %
    % |                                                                         | %
    % +-------------------------------------------------------------------------+ %         
    \subsection{Evaluation metric}\label{sec:metric}
        To evaluate retrieval performance, we use the \textit{mean Average Precision} (mAP) metric, as defined in~\citep{wu2007}, which captures the quality of video rankings. For each query, the \textit{Average  Precision} (AP) is calculated as
        \begin{equation}\label{eq3:map}
            AP = \frac{1}{n} \sum\limits_{i=0}^{n} \frac{i}{r_i},
        \end{equation}
        where $n$ is the number of relevant videos to the query video and $r_i$ is the rank of the $i$-th retrieved relevant video. The mAP is calculated by averaging the AP scores across all queries. Also, for the evaluation of the selector, we use the plot of mAP with respect to the total dataset percentage sent to the fine-grained student. The objective is to achieve high retrieval performance (in terms of mAP) with low dataset percentage. 
    
    % +-------------------------------------------------------------------------+ %
    % |                                                                         | %
    % |                       Implementation details                            | %
    % |                                                                         | %
    % +-------------------------------------------------------------------------+ %        
    \subsection{Implementation details}\label{sec:implementation_details}
        All of our models have been implemented with the PyTorch~\citep{paszke2019} library. For the teacher, we have re-implemented ViSiL~\citep{kordopatis2019b} following the same implementation details, i.e., for each video, we extracted 1 frame per second and used ResNet-50~\citep{he2016} for feature extraction using the output maps of the four residual blocks, resulting in $D=3840$. The PCA-whitening layer was learned from 1M region vectors sampled from VCDB. In all of our experiments, the weights of the feature extraction CNN and whitening layer remained fixed. We sampled 2000 triplets in each epoch. The teacher was trained for 200 epochs with 4 videos per batch using the raw video frames. We employed Adam optimization~\citep{kingma2014} with learning rate $10^{-5}$. Other parameters were set to $\gamma=0.5$, $r=0.1$ and $W=64$, similarly to~\citep{kordopatis2019b}.
        
        For the students, we used the same feature extraction process as in the teacher, and the same PCA-whitening layer was used for whitening and dimensionality reduction. We empirically set $D=512$ as the dimensions of the reduced region vectors. The students were trained with a batch size of 64 video pairs for 300 epochs, using only the extracted video features. Also, during training, we applied temporal augmentations, i.e., random frame drop, fast forward, and slow motion, with 0.1 probability each. We employed Adam optimization~\citep{kingma2014} with learning rate $10^{-5}$ and $10^{-4}$ for the course- and fine-grained students, respectively. For the fine-grained binarization student, the binarization layer was initialized with the ITQ algorithm~\citep{gong2012}, learned on 1M region vectors sampled from our dataset, as we observed better convergence than random initialization, and with $L=512$ bits. For the coarse-grained student's training, the teacher's similarities were rescaled to $[0,1]$ leading to better performance. Also, we used one layer in the transformer, with 8 heads for multi-head attention and 2048 dimension for the feed-forward network. For the NetVLAD module, we used 64 clusters, and a fully-connected layer with 1024 output dimensions and Layer Normalization~\citep{ba2016}. For the fine-grained students' training, we employed the similarity regularization loss from~\citep{kordopatis2019b}, weighted with $10^{-3}$, arriving at marginal performance improvements.
        
        For the selector, we used the same feature extraction scheme that was used for the students. It was trained with a batch size of 64 video pairs for 100 epochs, using only the extracted video features. At each epoch, we sampled 5,000 video pairs from each class. We employed Adam optimization~\citep{kingma2014} with learning rate $10^{-4}$. For the fully-connected layers of the MLP, we used 100 hidden units and 0.5 dropout rate. For the training of the selector model, the similarities of the fine-grained student were rescaled to $[0,1]$ to match similarities calculated from the coarse-grained student. Finally, we used a threshold of $t=0.2$ for the class separation, unless stated otherwise.

%%%%%%%%%%%%%%%%%%%%%%%%%%%%%%%%%%%%%%%%%%%%%%%%%%%%%%%%%%%%%%%%%%%%%%%%%%%%%%%%
%%                                                                            %%
%%                              [Experiments]                                 %%
%%                                                                            %%
%%%%%%%%%%%%%%%%%%%%%%%%%%%%%%%%%%%%%%%%%%%%%%%%%%%%%%%%%%%%%%%%%%%%%%%%%%%%%%%%
\section{Experiments}\label{sec:experiments}
    In this section, the experimental results of the proposed approach are provided. First, a comprehensive ablation study on the FIVR-5K dataset is presented, evaluating the proposed students and the overall approach under different configurations to gain better insight into its behaviour (Sect. \ref{sec:ablation_study}). Then, we compare the performance and requirements of the developed solutions against several methods from the literature on the four benchmark datasets (Sect. \ref{sec:soa_comparison}).

    % +-------------------------------------------------------------------------+ %
    % |                                                                         | %
    % |                           Ablation study                                | %
    % |                                                                         | %
    % +-------------------------------------------------------------------------+ %
    \subsection{Ablation study}\label{sec:ablation_study}
        
        % --------------------------------------------------------------------------    
        % Teacher                             : $\textbf{T}$
        % Fine-grained attention student      : $\textbf{S}^f_\mathcal{A}$
        % Fine-grained binarization student   : $\textbf{S}^f_\mathcal{B}$
        % Coarse-grained student              : $\textbf{S}^c$
        % --------------------------------------------------------------------------
        % $\textbf{S}^f_\mathcal{A}$, $\textbf{S}^f_\mathcal{B}$, and $\textbf{S}^c$

            \renewcommand{\arraystretch}{1.25}
            \begin{table}[t]
              \centering
              \caption{Comparison of the teacher $\textbf{T}$ and the students $\textbf{S}^f_\mathcal{A}$, $\textbf{S}^f_\mathcal{B}$, and $\textbf{S}^c$, in terms of mAP on FIVR-5K and computational requirements, i.e., storage space in KiloBytes (KB) per video and computational time in Seconds (Sec) per query.}
              \centering
              \begin{tabular}{|c|}
                \hline
                \textbf{Net.}              \\ \hline\hline
                  $\textbf{T}$                \\ \hline
                  $\textbf{S}^f_\mathcal{A}$  \\ \hline
                  $\textbf{S}^f_\mathcal{B}$  \\ \hline
                  $\textbf{S}^c$              \\ \hline
                \end{tabular}
              \begin{tabular}{|c|c|c|}
                \hline
                \textbf{DSVR}  &   \textbf{CSVR}   &  \textbf{ISVR}   \\ \hline\hline
                0.882          &           0.872   &          0.783   \\ \hline
                \textbf{0.893} &   \textbf{0.882}  &  \textbf{0.803}  \\ \hline
                0.879          &           0.868   &          0.788   \\ \hline
                0.634          &           0.647   &          0.608   \\ \hline
                \end{tabular}
              \begin{tabular}{|r|c|}
                \hline
                \multicolumn{1}{|c|}{\textbf{KB}}    &    \textbf{Sec}   \\ \hline\hline
                                           15124    &           10.10   \\ \hline
                                            2016    &           3.300   \\ \hline
                                              63    &           3.250   \\ \hline
                                               4    &           0.018   \\ \hline
                \end{tabular}
              \label{tab:retrieval_performance}
            \end{table}
        
        \subsubsection{Retrieval performance of the individual networks}\label{sec:retrieval_performance}
            In Table~\ref{tab:retrieval_performance}, we show the performance and storage/time requirements of the teacher $\textbf{T}$ and the three proposed student networks, namely, $\textbf{S}^f_\mathcal{A}$, $\textbf{S}^f_\mathcal{B}$, and $\textbf{S}^c$, trained with the proposed scheme. The fine-grained attention student $\textbf{S}^f_\mathcal{A}$ achieves the best results on all evaluation tasks, outperforming the teacher $\textbf{T}$ by a large margin. Also, the fine-grained binarization student $\textbf{S}^f_\mathcal{B}$ reports performance very close to the teacher's on the DSVR and CSVR tasks, and it outperforms the teacher on the ISVR task, using only quantized features with lower dimensionality than the ones used by the teacher and therefore requiring up to 240 times less storage space. This highlights the effectiveness of the proposed training scheme and the high quality of the collected dataset. Furthermore, both fine-grained students have similar time requirements, and they are three times faster than the teacher because they process lower dimensionality features. Finally, as expected, the coarse-grained student $\textbf{S}^c$ results in the worst performance compared to the other networks, but it has the lowest requirements in terms of both storage space and computational time.

            \begin{table}[t]
              \centering
              \caption{Comparison of the teacher $\textbf{T}$ and the students $\textbf{S}^f_\mathcal{A}$, $\textbf{S}^f_\mathcal{B}$, and $\textbf{S}^c$ trained on different datasets and training schemes in terms of mAP on FIVR-5K.}
              \scalebox{0.83}{
              \begin{tabular}{|c|c|c|c|c|}
                \hline
                  \multirow{2}{*}{\textbf{Task}} & \multirow{2}{*}{\textbf{Net.}} &   \textbf{VCDB}   &   \textbf{VCDB}   &   \textbf{DnS-100K}     \\ 
                  &  &   w/ supervision   &   w/ distillation   &   w/ distillation     \\ \hline\hline
                  \multirow{4}{*}{\textbf{DSVR}} 
                  & $\textbf{T}$               & 0.882 & - & - \\ \cline{2-5}
                  & $\textbf{S}^f_\mathcal{A}$ & 0.821 & 0.873 & 0.893 \\ \cline{2-5}
                  & $\textbf{S}^f_\mathcal{B}$ & 0.846 & 0.868 & 0.879 \\ \cline{2-5}
                  & $\textbf{S}^c$             & 0.444 & 0.510 & 0.634 \\ \hline\hline
                  \multirow{4}{*}{\textbf{CSVR}} 
                  & $\textbf{T}$               & 0.872 & - & - \\ \cline{2-5}
                  & $\textbf{S}^f_\mathcal{A}$ & 0.801 & 0.861 & 0.882 \\ \cline{2-5}
                  & $\textbf{S}^f_\mathcal{B}$ & 0.837 & 0.859 & 0.868 \\ \cline{2-5}
                  & $\textbf{S}^c$             & 0.443 & 0.520 & 0.647 \\ \hline\hline
                  \multirow{4}{*}{\textbf{ISVR}}
                  & $\textbf{T}$               & 0.783 & - & - \\ \cline{2-5}
                  & $\textbf{S}^f_\mathcal{A}$ & 0.727 & 0.787 & 0.803 \\ \cline{2-5}
                  & $\textbf{S}^f_\mathcal{B}$ & 0.769 & 0.781 & 0.788 \\ \cline{2-5}
                  & $\textbf{S}^c$             & 0.400 & 0.491 & 0.608 \\ \hline
                \end{tabular}
                }
              \label{tab:supervised_vs_distillation}
            \end{table}

        \subsubsection{Distillation vs Supervision}\label{sec:distillation_vs_supervision}
            
            In Table~\ref{tab:supervised_vs_distillation}, we show the performance of the teacher $\textbf{T}$ trained with supervision on VCDB (as proposed in~\citep{kordopatis2019b} and used for our teacher training) and the three proposed students, namely, $\textbf{S}^f_\mathcal{A}$, $\textbf{S}^f_\mathcal{B}$, and $\textbf{S}^c$, trained under various combinations: (i) with supervision on VCDB (same as the original teacher), (ii) with distillation on VCDB, and (iii) with distillation on the DnS-100K dataset. It is evident that the proposed training scheme using a large unlabelled dataset leads to considerably better retrieval performance compared to the other setups for all students. Also, it is noteworthy that training students with supervision, same as the teacher, results in a considerable drop in performance compared to distillation on either dataset. The students achieve better results when trained with DnS-100K instead of VCDB. An explanation for this is that our dataset contains various video relations (not only near-duplicates as in VCDB) and represents a very broad and diverse domain (by contrast to VCDB, which consists of randomly selected videos), resulting in better retrieval performance for the students.
    
        \begin{table}[t]
              \centering
              \caption{Comparison of students $\textbf{S}^f_\mathcal{A}$, $\textbf{S}^f_\mathcal{B}$, and $\textbf{S}^c$, in terms of mAP, trained with different amount of training data on FIVR-5K.}
              \begin{tabular}{|c|c|c|c|c|c|}
                  \hline
                  \multirow{2}{*}{\textbf{Task}} & \multirow{2}{*}{\textbf{Student}} &
                  \multicolumn{4}{c|}{\textbf{\textbf{Dataset (\%)}}} \\ \cline{3-6}
                  & &   \textbf{25}   &   \textbf{50}   &   \textbf{75}  &  \textbf{100} \\\hline\hline
                  \multirow{3}{*}{\textbf{DSVR}} 
                  & $\textbf{S}^f_\mathcal{A}$ & 0.888 & \textbf{0.894} & 0.893 & 0.893  \\ \cline{2-6}
                  & $\textbf{S}^f_\mathcal{B}$ & 0.868 & 0.871 & 0.873 & \textbf{0.879}  \\ \cline{2-6}
                  & $\textbf{S}^c$             & 0.573 & 0.608 & 0.621 & \textbf{0.634}  \\ \hline\hline
                  \multirow{3}{*}{\textbf{CSVR}} 
                  & $\textbf{S}^f_\mathcal{A}$ & 0.877 & \textbf{0.883} & 0.882 & \textbf{0.883}  \\ \cline{2-6}
                  & $\textbf{S}^f_\mathcal{B}$ & 0.858 & 0.861 & 0.863 & \textbf{0.868}  \\ \cline{2-6}
                  & $\textbf{S}^c$             & 0.586 & 0.621 & 0.633 & \textbf{0.647}  \\ \hline\hline
                  \multirow{3}{*}{\textbf{ISVR}} 
                  & $\textbf{S}^f_\mathcal{A}$ & 0.801 & 0.803 & 0.803 & \textbf{0.805}  \\ \cline{2-6}
                  & $\textbf{S}^f_\mathcal{B}$ & 0.776 & 0.779 & 0.783 & \textbf{0.788}  \\ \cline{2-6}
                  & $\textbf{S}^c$             & 0.564 & 0.583 & 0.594 & \textbf{0.608}  \\ \hline
                \end{tabular}
              \label{tab:dataset_percentage}
            \end{table}
        
        \subsubsection{Impact of dataset size}\label{sec:dataset_size}
            
            In Table~\ref{tab:dataset_percentage}, we show the performance of the proposed students, namely, $\textbf{S}^f_\mathcal{A}$, $\textbf{S}^f_\mathcal{B}$, and $\textbf{S}^c$, in terms of mAP, when they are trained with different percentages of the collected DnS-100K dataset (that is, 25\%, 50\%, 75\%, and 100\%). We report large differences in performance for the fine-grained binarization student $\textbf{S}^f_\mathcal{B}$ and the coarse-grained student $\textbf{S}^c$. We note that the more data is used for training, the better their retrieval results are. On the other hand, the fine-grained attention student's $\textbf{S}^f_\mathcal{A}$ performance remains relatively steady, regardless of the amount used for training. We attribute this behaviour to the fact that $\textbf{S}^f_\mathcal{A}$ learns to weigh the input features without transforming them; hence, a smaller dataset with real video pairs with diverse relations, as in our collected dataset, is adequate for its robust performance.

        \subsubsection{Student performance with different teachers}\label{sec:differnt_teacher}
            
            In Table~\ref{tab:different_teachers}, we show the performance of the proposed students, namely, $\textbf{S}^f_\mathcal{A}$, $\textbf{S}^f_\mathcal{B}$, and $\textbf{S}^c$, in terms of mAP, when they are trained/distilled using different teachers. More specifically, using as a teacher: (i) the original teacher $\textbf{T}$, leading to the student $\textbf{S}^{f(1)}_\mathcal{A}$, (ii) the fine-grained attention student $\textbf{S}^{f(1)}_\mathcal{A}$, leading to the student $\textbf{S}^{f(2)}_\mathcal{A}$ (first iteration), and (iii) the fine-grained attention student $\textbf{S}^{f(2)}_\mathcal{A}$ (second iteration). In the case of fine-grained students, training with the $\textbf{S}^{f(1)}_\mathcal{A}$ and $\textbf{S}^{f(2)}_\mathcal{A}$ yields large performance boost in comparison to original teacher $\textbf{T}$. More precisely, the fine-grained attention student $\textbf{S}^{f}_\mathcal{A}$ exhibits a total improvement of about 0.006 mAP comparing its results trained with the teacher $\textbf{T}$ (i.e., 0.893 mAP on DSVR task) and the $\textbf{S}^{f(2)}_\mathcal{A}$ (i.e., 0.899 mAP on DSVR task). A very considerable improvement has the fine-grained binarization student, i.e., training with $\textbf{S}^{f(1)}_\mathcal{A}$ gives a performance increase of almost 0.01 mAP on DSVR task, which further improves when trained with $\textbf{S}^{f(2)}_\mathcal{A}$ by 0.007. On the other hand, using a better teacher does not improve the performance of the coarse-grained student $\textbf{S}^c$.
            
            \begin{table}[t]
              \centering
              \caption{Comparison of students $\textbf{S}^f_\mathcal{A}$, $\textbf{S}^f_\mathcal{B}$, and $\textbf{S}^c$, in terms of mAP, trained with different teachers on FIVR-5K.}
              \renewcommand{\arraystretch}{1.2}
              \begin{tabular}{|c|c|c|c|c|c|}
                \hline
                \multirow{2}{*}{\textbf{Task}} & \multirow{2}{*}{\textbf{Student}} & \multicolumn{3}{c|}{\textbf{Teacher}} \\ \hhline{~~---}
                & & \textbf{$\textbf{T}$}   &   \textbf{$\textbf{S}^{f (1)}_\mathcal{A}$}   &   \textbf{$\textbf{S}^{f (2)}_\mathcal{A}$}     \\ \hline\hline
                  \multirow{3}{*}{\textbf{DSVR}} 
                  & $\textbf{S}^f_\mathcal{A}$    & 0.893 & 0.896 & \textbf{0.899} \\ \hhline{~----}
                  & $\textbf{S}^f_\mathcal{B}$    & 0.879 & 0.888 & \textbf{0.895} \\ \hhline{~----}
                  & $\textbf{S}^c$                & \textbf{0.634} & 0.631 & 0.632 \\ \hline\hline
                  \multirow{3}{*}{\textbf{CSVR}}  
                  & $\textbf{S}^f_\mathcal{A}$    & 0.882 & 0.884 & \textbf{0.887} \\ \hhline{~----}
                  & $\textbf{S}^f_\mathcal{B}$    & 0.868 & 0.878 & \textbf{0.883} \\ \hhline{~----}
                  & $\textbf{S}^c$                & \textbf{0.647} & 0.643 & 0.645 \\ \hline\hline
                  \multirow{3}{*}{\textbf{ISVR}}  
                  & $\textbf{S}^f_\mathcal{A}$    & 0.803 & 0.807 & \textbf{0.810} \\ \hhline{~----}
                  & $\textbf{S}^f_\mathcal{B}$    & 0.788 & 0.797 & \textbf{0.804} \\ \hhline{~----}
                  & $\textbf{S}^c$                & \textbf{0.608} & 0.603 & 0.607 \\ \hline
                \end{tabular}
              \label{tab:different_teachers}
            \end{table}

        \subsubsection{Student performance with different settings}\label{sec:different_settings}
            
            In this section, the retrieval performance of the proposed students is evaluated under different design choices.
            
            \textbf{Fine-grained attention student:} In Table~\ref{tab:attention_comparison}, we show how the adopted attention scheme ($\ell^2$-attention -- Sect.~\ref{sec:baseline_teacher}, or $h$-attention -- Sect.~\ref{sec:attention_student}) affects the performance of the student $\textbf{S}^f_\mathcal{A}$. Using $h$-attention leads to considerably better results compared to the $\ell^2$-attention, that was originally used in ViSiL~\citep{kordopatis2019b}.
            
            \begin{table}[t]
              \centering
              \caption{Comparison of different attention schemes of the fine-grained attention student $\textbf{S}^f_\mathcal{A}$, in terms of mAP on FIVR-5K.}
              \scalebox{0.92}{
              \begin{tabular}{|l|c|c|c|}
                \hline
                \textbf{attention} &   \textbf{DSVR}   &   \textbf{CSVR}   &   \textbf{ISVR}     \\ \hline\hline
                  $\ell^2$-attention      & \multirow{2}{*}{0.888} & \multirow{2}{*}{0.873} & \multirow{2}{*}{0.787} \\ 
                  \citep{kordopatis2019b}  & & & \\ \hline
                  $h$-attention           & \multirow{2}{*}{\textbf{0.893}} & \multirow{2}{*}{\textbf{0.882}} & \multirow{2}{*}{\textbf{0.803}} \\ 
                  \citep{yang2016}         & & & \\ \hline
                \end{tabular}
                }
              \label{tab:attention_comparison}
            \end{table}
            
            \textbf{Fine-grained binarization student:} In Table~\ref{tab:binarization_comparison}, we report the retrieval results of the fine-grained binarization student $\textbf{S}^f_\mathcal{B}$ implemented with different activation functions in the binarization layer, i.e., $\text{sgn}(\textbf{x})$ which is not differentiable so the layer weights remain fixed, $\text{tanh}(\beta \textbf{x})$, as proposed in~\citep{cao2017} with $\beta=10^3$, and the proposed $\EX[\text{sgn}(\textbf{x})]$ (Sect.~\ref{sec:binarization_student}). The binarization student with the proposed function achieves notably better results on all tasks, especially on ISVR. Moreover, we experimented with different number of bits for the region vectors and report results in Table~\ref{tab:binarization_bits}. As expected, larger region hash codes lead to better retrieval performance. Nevertheless, the student achieves competitive retrieval performance even with low number of bits per region vector.
            
            \begin{table}[t]
              \centering
              \caption{Comparison of different activation functions for fine-grained binarization student $\textbf{S}^f_\mathcal{B}$, in terms of mAP on FIVR-5K.}
              \begin{tabular}{|l|c|c|c|}
                \hline
                \textbf{activation} &   \textbf{DSVR}   &   \textbf{CSVR}   &   \textbf{ISVR}     \\ \hline\hline
                  $\text{sgn}(\textbf{x})$                        & 0.876   &  0.863   &  0.776  \\ \hline
                  $\text{tanh}(\beta \textbf{x})$~\citep{cao2017} & 0.875   &  0.861   &  0.781  \\ \hline
                  $\EX[\text{sgn}(\textbf{x})]$ (Ours)            & \textbf{0.879}   &  \textbf{0.868}   &  \textbf{0.788}  \\ \hline
                \end{tabular}
              \label{tab:binarization_comparison}
            \end{table}
            
            \begin{table}[t]
              \centering
              \caption{Comparison of different fine-grained binarization student $\textbf{S}^f_\mathcal{B}$ implemented with different number of bits per region vector, in terms of mAP on FIVR-5K.}
              \begin{tabular}{|c|c|c|c|}
                \hline
                \textbf{bits} &   \textbf{DSVR}   &   \textbf{CSVR}   &   \textbf{ISVR}     \\ \hline\hline
                  64    & 0.845   &  0.835   &  0.748  \\ \hline
                  128   & 0.862   &  0.849   &  0.766  \\ \hline
                  256   & 0.870   &  0.857   &  0.779  \\ \hline
                  512   & \textbf{0.879}   &  \textbf{0.868}   &  \textbf{0.788}  \\ \hline
                \end{tabular}
              \label{tab:binarization_bits}
            \end{table}
            
            \textbf{Coarse-grained student:} In Table~\ref{tab:global_comparison}, we report the performance of the coarse-grained student $\textbf{S}^c$ implemented under various combinations of its components. The proposed setup with all three components achieves the best results compared to the other configurations. The single component that provides the best results is the transformer network, followed by NetVLAD. Also, the attention mechanism provides a considerable boost in performance when applied. The second-best performance is achieved with the combination of the transformer module with the NetVLAD.
        
        \subsubsection{Selector network performance}\label{sec:selector_performance}
        
\begin{figure}[t]
                \centering
                \subfigure[DSVR]{\includegraphics[width=2.7cm]{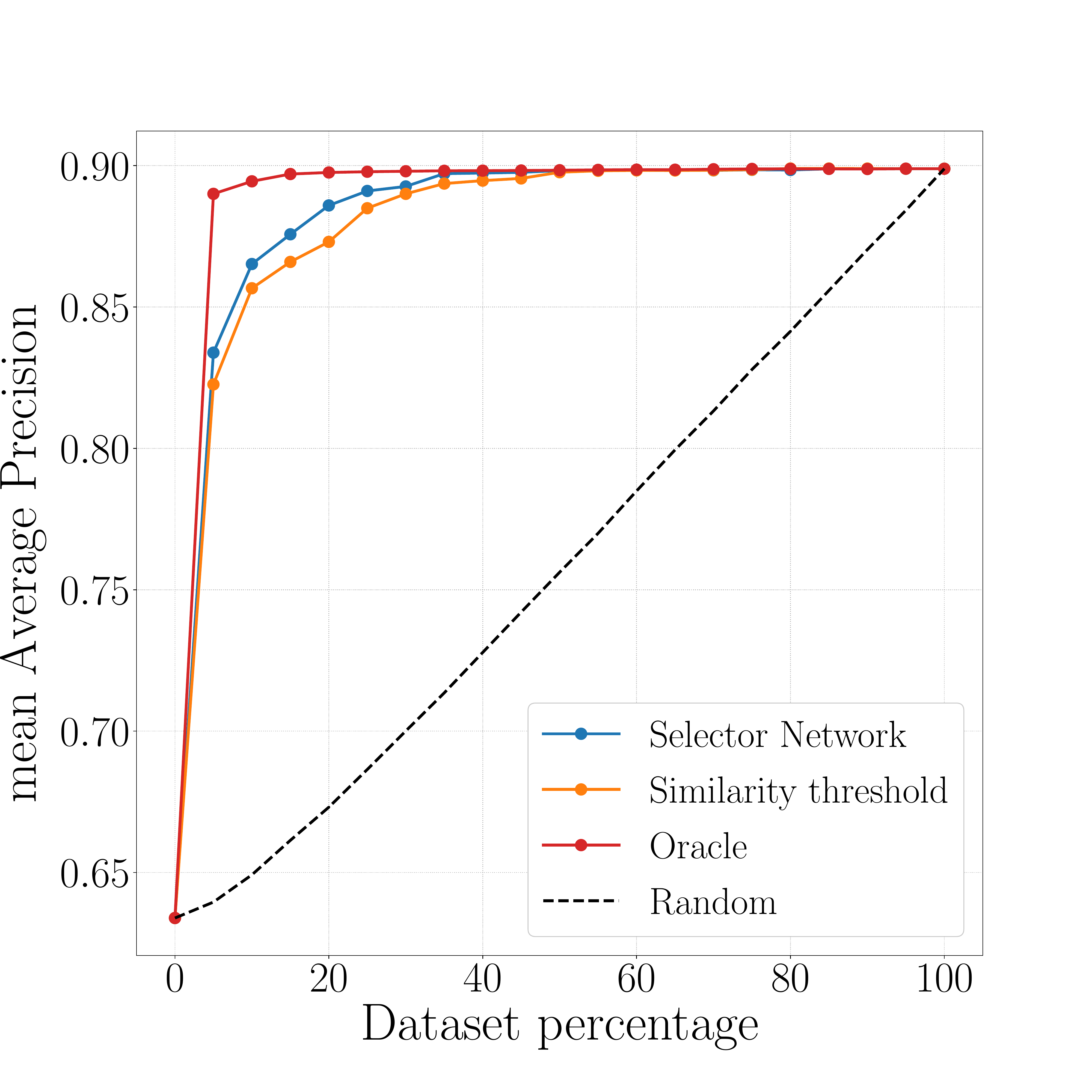}}
                \subfigure[CSVR]{\includegraphics[width=2.7cm]{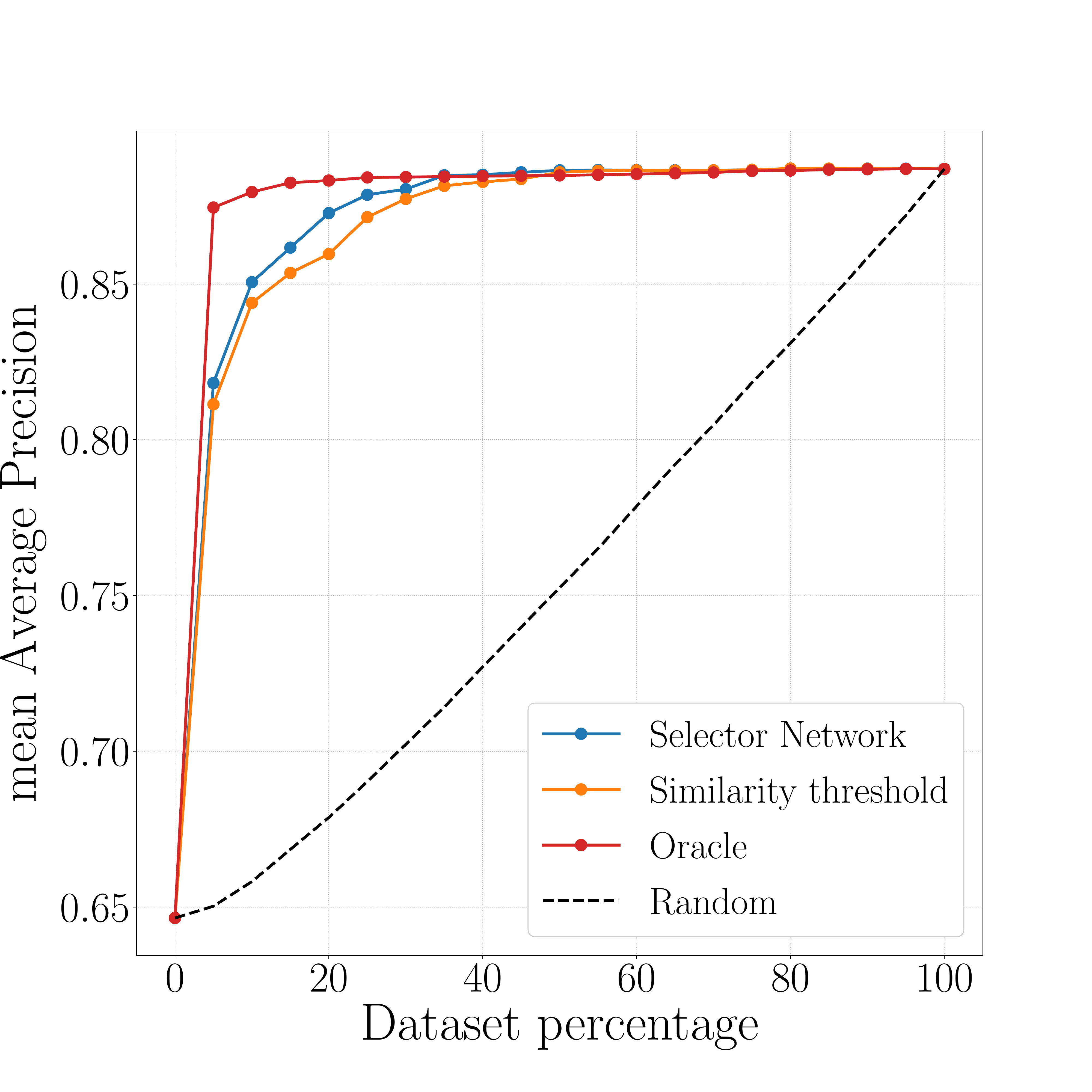}}
                \subfigure[ISVR]{\includegraphics[width=2.7cm]{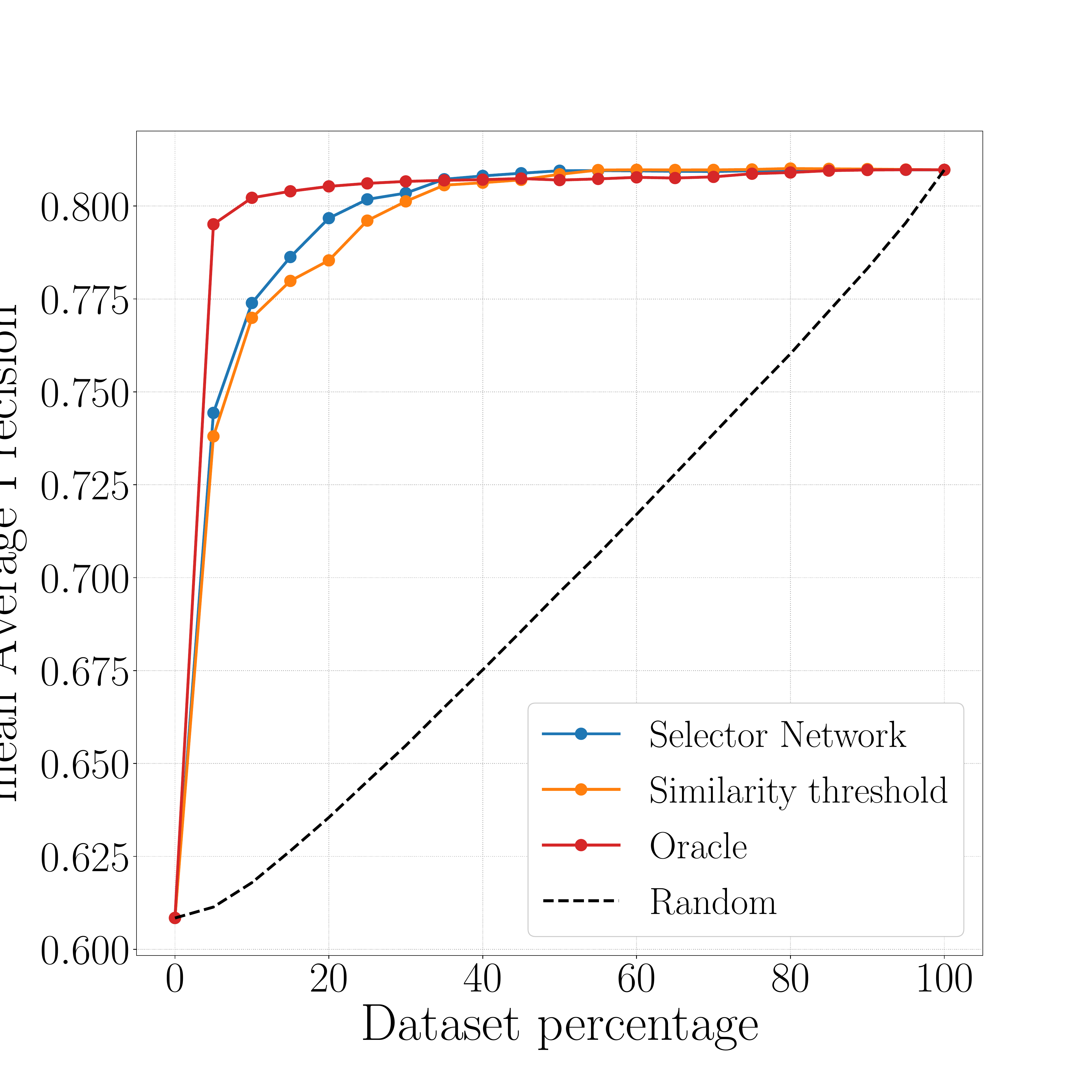}}
                \caption{mAP with respect to the dataset percentage sent to the fine-grained student for re-ranking based on four selectors: (i) the proposed selector network, (ii) a selector with naive similarity thresholding, (iii) an oracle selector, ranking videos based on the similarity difference between the two students, and (iv) a random selector.}
                \label{fig:selector_similarity_comparison}
            \end{figure}

            In this section, the performance of the proposed selector network is evaluated in comparison with the following approaches: (i) a selection mechanism that applies naive similarity thresholding for choosing between the coarse-grained and the fine-grained student, (ii) an oracle selector, where the similarity difference between the fine-grained and coarse-grained student is known and used for the re-ranking of video pairs, and (iii) a random selector that sends with a fixed probability videos to either the coarse-grained or the fine-grained student. Fig.~\ref{fig:selector_similarity_comparison} illustrates the performance of the DnS approach in terms of mAP with respect to the percentage of video pairs from the evaluation dataset sent to the fine-grained student. We consider that the closer the curves are to the upper left corner, the better their performance. For this experiment, we used the proposed fine-grained attention student $\textbf{S}^f_\mathcal{A}$ and the coarse-grained student $\textbf{S}^c$. All three runs outperform the performance of the random selector by a large margin on all dataset percentages. The oracle selector performs the best with considerable margin, highlighting that using the similarity difference between the two students (Sect.~\ref{sec:selector_training}) is a good optimization criterion. Furthermore, the proposed selector network outperforms the one with similarity thresholding on all tasks and percentages, i.e., in lower dataset percentages ($<25\%$) with a large margin. It achieves more than 0.85 mAP on the DSVR task with only 10\% of the video pairs in FIVR-5k sent to the fine-grained student.
        
            \begin{table}[t]
              \centering
              \caption{Comparison of different design choices of our coarse-grained student $\textbf{S}^c$, in terms of mAP on FIVR-5K.}
              \scalebox{0.90}{
              \begin{tabular}{|c|c|c|c|c|c|}
                \hline
                \textbf{Att.} & \textbf{Trans.} & \textbf{NetVLAD} &   \textbf{DSVR}   &   \textbf{CSVR}   &   \textbf{ISVR}     \\ \hline\hline
                  \checkmark & - & -                   & 0.595   &  0.600   &  0.564  \\ \hline
                  - & \checkmark & -                   & 0.612   &  0.622   &  0.590  \\ \hline
                  - & - & \checkmark                   & 0.600   &  0.610   &  0.578  \\ \hline
                  \checkmark & \checkmark & -          & 0.620   &  0.628   &  0.591  \\ \hline
                  \checkmark & - & \checkmark          & 0.609   &  0.618   &  0.584  \\ \hline
                  - & \checkmark & \checkmark          & 0.630   &  0.637   &  0.600  \\ \hline
                  \checkmark & \checkmark & \checkmark & \textbf{0.634}   &  \textbf{0.647}   &  \textbf{0.608}  \\ \hline
                \end{tabular}
                }
              \label{tab:global_comparison}
            \end{table}

        \subsubsection{Impact of threshold on the selector performance}\label{sec:threshold_impact}
            
\begin{figure}[t]
            \centering
            \subfigure[DSVR]{\includegraphics[width=2.7cm]{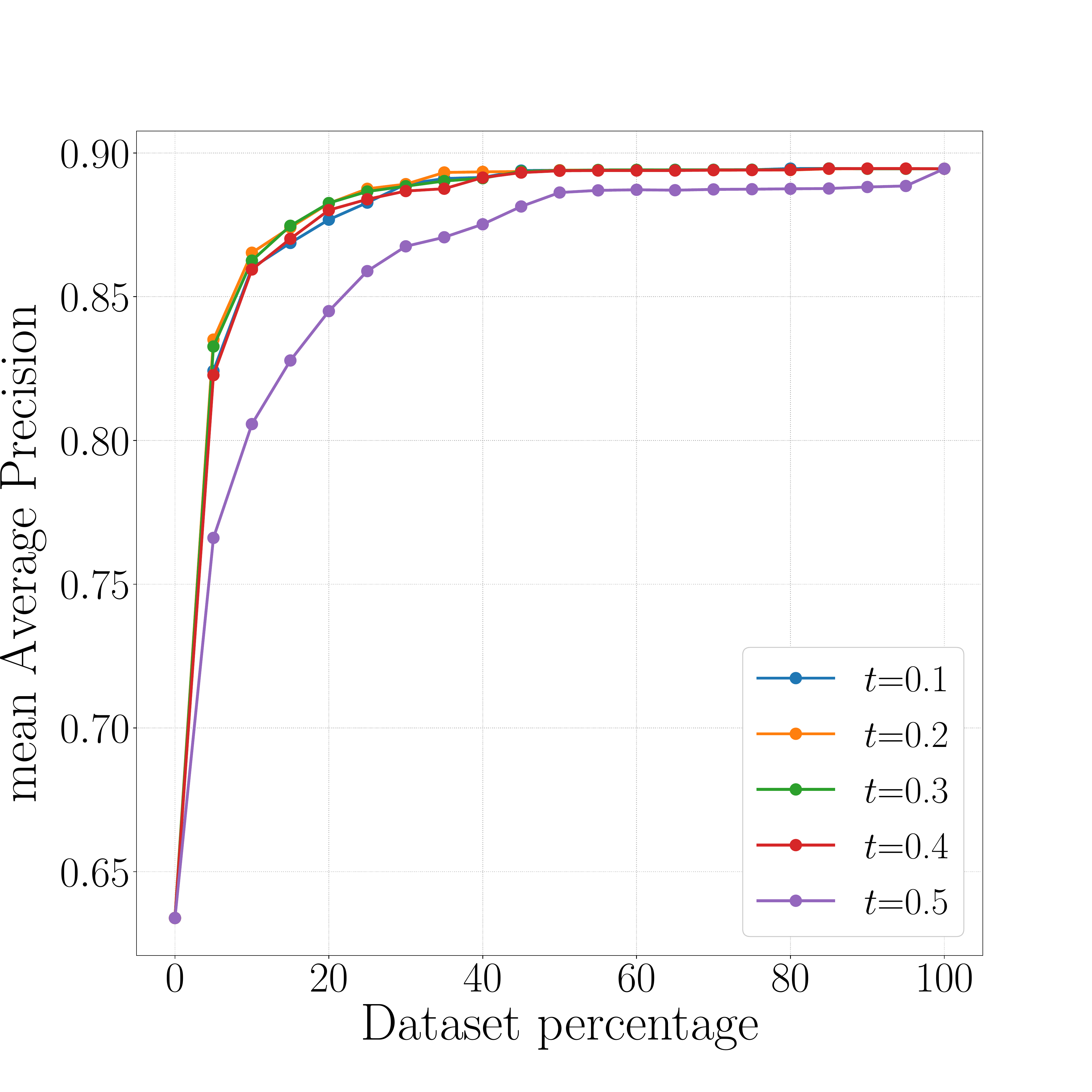}}
            \subfigure[CSVR]{\includegraphics[width=2.7cm]{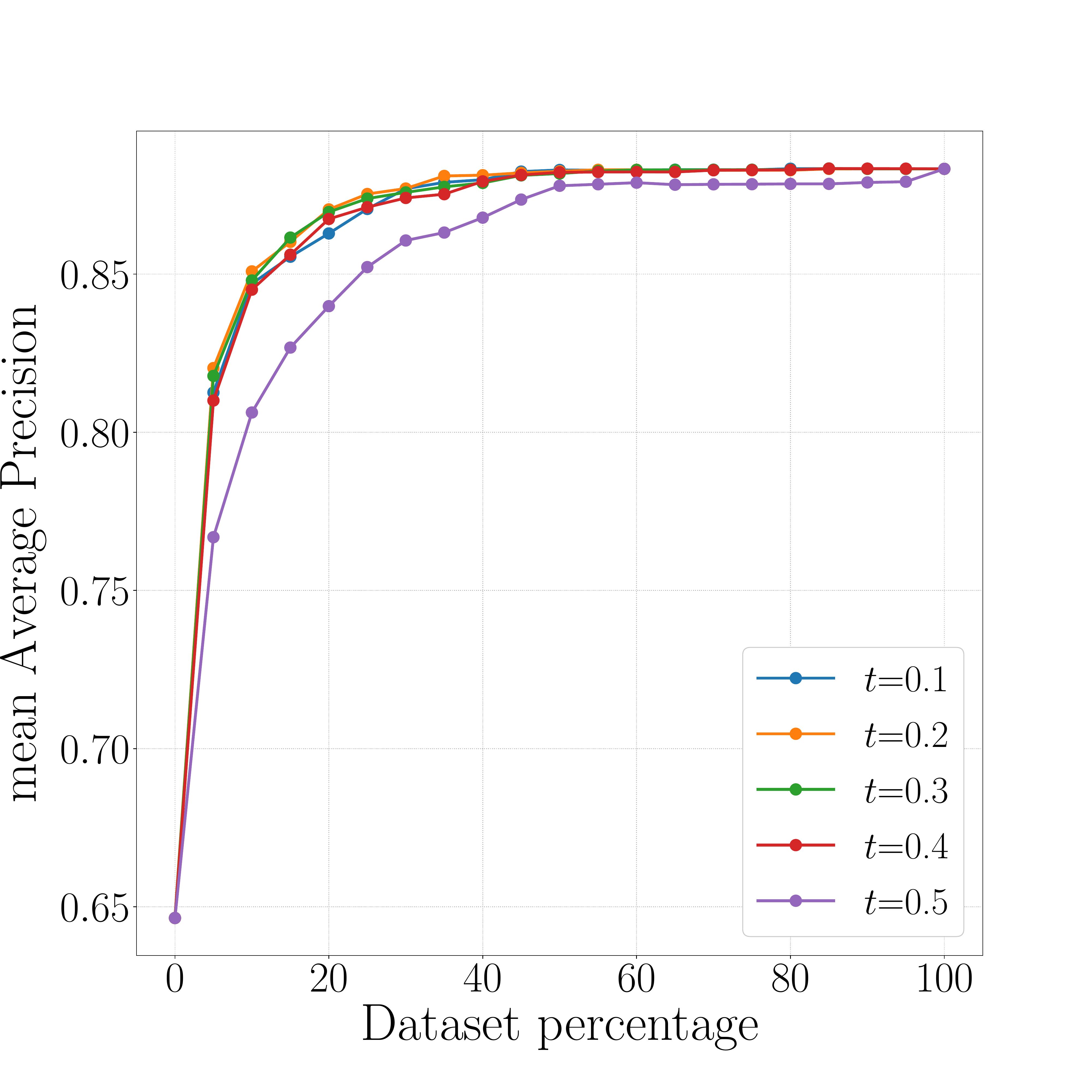}}
            \subfigure[ISVR]{\includegraphics[width=2.7cm]{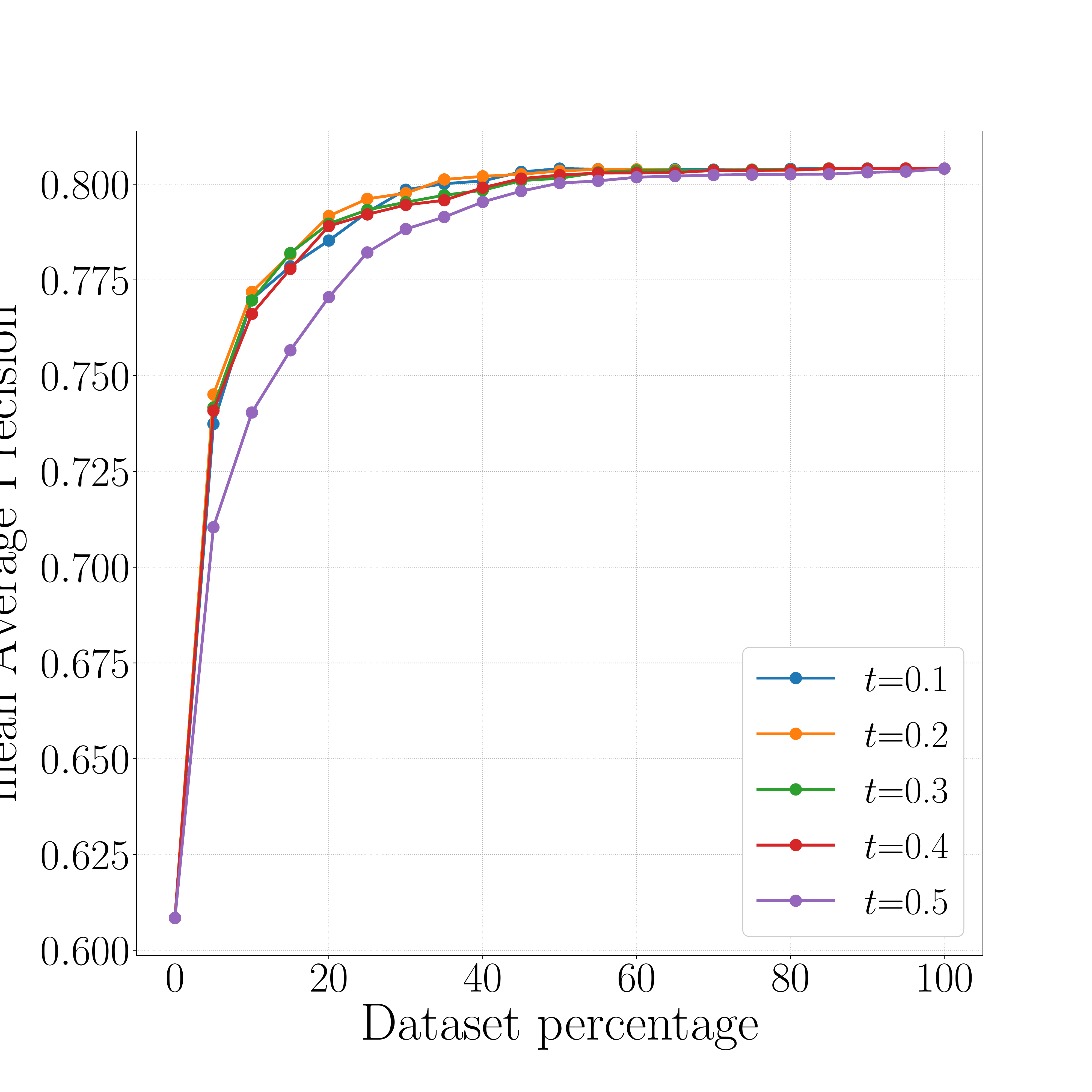}}
            \caption{mAP with respect to the dataset percentage sent to the fine-grained student for re-ranking based on our selector network trained with different values for the threshold $t$.}
            \label{fig:selector_threshold}
        \end{figure}

            In this section, we assess the impact of the threshold parameter $t$ that is used to obtain binary labels for the selector network (see Sect.~\ref{sec:selector_training}, equation (\ref{eq:label})), on the retrieval performance. To do so, we report the mAP as a function of the dataset percentage sent to the fine-grained student for re-ranking -- we do so for selectors trained with different values of $t$ in order to compare the curves. The results are shown in Fig.~\ref{fig:selector_threshold}. The best results are obtained for $t=0.2$; however, the performance is rather stable for thresholds between 0.1 and 0.4, as well. For threshold values $>0.4$, the performance drops considerably on all evaluation tasks.
    
        \begin{table*}[t]
          \centering
          \renewcommand{\arraystretch}{1.2}
          \caption {mAP comparison of our proposed students and re-ranking method against several video retrieval methods on four evaluation datasets. $\ssymbol{2}$ indicates that the runs are implemented with the same features extracted with the same process as ours. $\ssymbol{1}$ indicates that the corresponding results are on different dataset split.}
          \scalebox{0.82}{
          \begin{tabular}{|l|l|c|c|c|c|c|c|c|c|c|}
            \hline 
            & \multirow{2}{*}{\textbf{Approach}} & \multicolumn{3}{c|}{\textbf{FIVR-200K}} & \multicolumn{4}{c|}{\textbf{CC\_WEB\_VIDEO}} & \multirow{2}{*}{\textbf{SVD}} & \multirow{2}{*}{\textbf{EVVE}} \\ \cline{3-9}
            & &  \textbf{DSVR}   & \textbf{CSVR}    & \textbf{ISVR} & \textbf{cc\_web} & \textbf{cc\_web}$^*$ & \textbf{cc\_web}$_c$ & \textbf{cc\_web}$^*_c$ & & \\ \hline\hline
            \multirow{10}{*}{\rotatebox[origin=c]{90}{\centering \textbf{Coarse-grained}}}
            &\textbf{ITQ}$\ssymbol{2}$~\citep{gong2012}        & 0.491 & 0.472 & 0.402 & 0.967 & 0.941 & 0.976 & 0.954 & 0.842 & 0.606 \\ \cline{2-11}
            &\textbf{MFH}$\ssymbol{2}$~\citep{song2013}        & 0.525 & 0.507 & 0.424 & 0.972 & 0.950 & 0.981 & 0.967 & 0.849 & 0.604 \\ \cline{2-11}
            &\textbf{CSQ}~\citep{yuan2020}                     & 0.267 & 0.252 & 0.219 & 0.948 & 0.899 & 0.954 & 0.909 & 0.364 & 0.400 \\ \cline{2-11}
            &\textbf{BoW}$\ssymbol{2}$~\citep{cai2011}         & 0.699 & \textbf{0.674} & \textbf{0.581} & 0.975 & 0.958 & \textbf{0.985} & \textbf{0.977} & 0.568 & 0.539 \\ \cline{2-11}
            &\textbf{LBoW}~\citep{kordopatis2017a}             & \textbf{0.700} & 0.666 & 0.566 & \textbf{0.976} & \textbf{0.960} & 0.984 & 0.975 & 0.756 & 0.500 \\ \cline{2-11}
            &\textbf{DML}~\citep{kordopatis2017b}              & 0.411 & 0.392 & 0.321 & 0.971 & 0.941 & 0.979 & 0.959 & 0.785 & 0.531 \\ \cline{2-11}
            &\textbf{DML}$\ssymbol{2}$~\citep{kordopatis2017b} & 0.503 & 0.487 & 0.410 & 0.971 & 0.951 & 0.979 & 0.965 & 0.850 & 0.611 \\ \cline{2-11}
            &\textbf{R-UTS-GV}\citep{liang2019}               & 0.509 & 0.498 & 0.432 & - & - & - & - & - & -   \\ \cline{2-11}
            &\textbf{TCA$_c$}$\ssymbol{2}$~\citep{shao2021}    & 0.570 & 0.553 & 0.473 & 0.973 & 0.947 & 0.983 & 0.965 & - & 0.598$\ssymbol{1}$ \\ \cline{2-11}
            &\textbf{Coarse-grained student-}$\textbf{S}^c$ (Ours)                           & 0.574 & 0.558 & 0.476 & 0.972 & 0.952 & 0.980 & 0.967 & \textbf{0.868} & \textbf{0.636} \\ \hline\hline
            \multirow{11}{*}{\rotatebox[origin=c]{90}{\centering \textbf{Fine-grained}}}  
            &\textbf{TMK}$\ssymbol{2}$~\citep{poullot2015}     & 0.524 & 0.507 & 0.425 & 0.977 & 0.959 & 0.986 & 0.975 & 0.863 & 0.618 \\ \cline{2-11}
            &\textbf{LAMV}~\citep{baraldi2018}                 & 0.496 & 0.466 & 0.371 & 0.975 & 0.956 & 0.986 & 0.975 & 0.781 & 0.531 \\ \cline{2-11}
            &\textbf{LAMV}$\ssymbol{2}$~\citep{baraldi2018}    & 0.619 & 0.587 & 0.479 & 0.978 & 0.964 & 0.988 & 0.982 & 0.880 & 0.620 \\ \cline{2-11}
            &\textbf{TN}$\ssymbol{2}$~\citep{tan2009}          & 0.844 & 0.804 & 0.660 & 0.982 & 0.970 & 0.993 & 0.989 & 0.894 & 0.471 \\ \cline{2-11}
            &\textbf{DP}$\ssymbol{2}$~\citep{chou2015}         & 0.827 & 0.783 & 0.642 & 0.980 & 0.966 & 0.991 & 0.987 & 0.880 & 0.580 \\ \cline{2-11}
            &\textbf{R-UTS-FRP}~\citep{liang2019}              & 0.769 & 0.724 & 0.611 & - & - & - & - & - & - \\ \cline{2-11}
            &\textbf{A-DML}~\citep{wang2021}                   & 0.627 & - & - & 0.964 & 0.949 & - & - & 0.885$\ssymbol{1}$ & - \\ \cline{2-11}
            &\textbf{TCA$_f$}$\ssymbol{2}$~\citep{shao2021}    & 0.877 & 0.830 & 0.703 & 0.983 & 0.969 & 0.994 & 0.990 & - & 0.603$\ssymbol{1}$ \\ \cline{2-11}
            &\textbf{ViSiL}~\citep{kordopatis2019b}            & 0.899 & 0.854 & 0.723 & \textbf{0.985} & 0.973 & \textbf{0.995} & \textbf{0.992} & 0.881 & \textbf{0.658} \\ \cline{2-11}
            &\textbf{Fine-grained att. student-}$\textbf{S}^f_\mathcal{A}$ (Ours)   & \textbf{0.921} & \textbf{0.875} & \textbf{0.741} & 0.984 & 0.973 & \textbf{0.995} & \textbf{0.992} & \textbf{0.902}& 0.651 \\ \cline{2-11}
            &\textbf{Fine-grained bin. student-}$\textbf{S}^f_\mathcal{B}$ (Ours) & 0.909 & 0.863 & 0.729 & 0.984 & \textbf{0.974} & \textbf{0.995} & 0.991 & 0.891 & 0.640 \\ \hline\hline
            \multirow{8}{*}{\rotatebox[origin=c]{90}{\centering \textbf{Re-ranking}}}
            &\textbf{PPT}~\citep{chou2015}                     & - & - & - & 0.959 & - & - & - & - & - \\ \cline{2-11}
            &\textbf{HM}~\citep{liang2020}                     & - & - & - & 0.977 & - & - & - & - & - \\ \cline{2-11}
            &\textbf{TMK$\ssymbol{2}$+QE}~\citep{poullot2015}  & 0.580 & 0.564 & 0.480 & 0.977 & 0.960 & 0.986 & 0.976 & 0.774 & 0.648 \\ \cline{2-11}
            &\textbf{LAMV$\ssymbol{2}$+QE}~\citep{baraldi2018} & 0.659 & 0.629 & 0.520 & \textbf{0.979} & \textbf{0.964} & \textbf{0.990} & \textbf{0.984} & 0.786 & \textbf{0.653} \\ \cline{2-11}
            &$\textbf{DnS}^{5\%}_\mathcal{A}$ (Ours)          & 0.874 & 0.829 & 0.699 & 0.972 & 0.951 & 0.983 & 0.969 & 0.895 & 0.594 \\ \cline{2-11} 
            &$\textbf{DnS}^{5\%}_\mathcal{B}$ (Ours)          & 0.862 & 0.817 & 0.687 & 0.970 & 0.948 & 0.981 & 0.967 & 0.884 & 0.584 \\ \cline{2-11}
            &$\textbf{DnS}^{30\%}_\mathcal{A}$ (Ours)         & \textbf{0.913} & \textbf{0.868} & \textbf{0.733} & 0.978 & 0.958 & \textbf{0.990} & 0.977 & \textbf{0.902} & 0.646 \\ \cline{2-11}
            &$\textbf{DnS}^{30\%}_\mathcal{B}$ (Ours)         & 0.900 & 0.854 & 0.720 & 0.977 & 0.954 & 0.988 & 0.974 & 0.894 & 0.634 \\ \hline
          \end{tabular}
          }
          \label{tab:mAP_comparison}
        \end{table*}
    
    % +-------------------------------------------------------------------------+ %
    % |                                                                         | %
    % |                   Comparison against state-of-the-art                   | %
    % |                                                                         | %
    % +-------------------------------------------------------------------------+ %
    \subsection{Comparison with State of the Art}\label{sec:soa_comparison}
        
        In this section, the proposed approach is compared with several methods from the literature on four datasets. In all experiments, the fine-grained attention student $\textbf{S}^{f(2)}_\mathcal{A}$ is used as the teacher. We report the results of our re-ranking DnS scheme using both fine-grained students and sending the 5\% and 30\% of the dataset videos per query for re-ranking based on our selector score. We compare its performance with several coarse-grained, fine-grained, and re-ranking approaches: \textbf{ITQ}~\citep{gong2012} and \textbf{MFH}~\citep{song2013} are two unsupervised and \textbf{CSQ}~\citep{yuan2020} a supervised video hashing methods using Hamming distance for video ranking, \textbf{BoW} \citep{cai2011} and \textbf{LBoW}~\citep{kordopatis2017a} extracts video representations based on BoW schemes with tf-idf weighting, \textbf{DML}~\citep{kordopatis2017b} extract a video embedding based on a network trained with DML, \textbf{R-UTS-GV} and \textbf{R-UTS-FRP}~\citep{liang2019} are a coarse- and fine-grained methods trained with a teacher-student setup distilling feature representations, \textbf{TCA$_c$} and \textbf{TCA$_f$} \citep{shao2021} are a coarse- and fine-grained methods using a transformer-based architecture trained with contrastive learning, \textbf{TMK}~\citep{poullot2015} and \textbf{LAMV}~\citep{baraldi2018} extracts spatio-temporal video representations based on Fourier transform, which are also combined with \textbf{QE}~\citep{douze2013}, \textbf{TN}~\citep{tan2009} employs a temporal network to find video segments with large similarity, \textbf{DP}~\citep{chou2015} is a dynamic programming scheme for similarity calculation, \textbf{A-DML}~\citep{wang2021} assess video similarity extracting multiple video representations based on a multi-head attention network, \textbf{PPT}~\citep{chou2015} is a re-ranking method with a BoW-based indexing scheme combined with DP for reranking, and \textbf{HM}~\citep{liang2020} is also a re-ranking method using a concept-based similarity and a BoW-based method for refinement, and our re-implementation of \textbf{ViSiL}~\citep{kordopatis2019b}. From the aforementioned methods, we have re-implemented BoW, TN, and DP, and we use the publicly available implementations for ITQ, MFH, CSQ, DML, TMK, and LAMV. For the rest, we provide the results reported in the original papers. Also, for fair comparison, we have implemented (if possible) the publicly available methods using our extracted features.

        \begin{table*}[t]
          \centering
          \renewcommand{\arraystretch}{1.3}
          \caption {Performance in mAP, storage in KiloBytes (KB) and time in Seconds (Sec) requirements of our proposed students and re-ranking method and several video retrieval implemented with the same features. $\ssymbol{1}$ indicates that the corresponding results are on different dataset split.}
          \scalebox{0.75}{
          \begin{tabular}{|l|l|c|r|c|c|r|c|c|r|c|c|r|c|c|r|c|}
            \hline 
            & \multirow{2}{*}{\textbf{Approach}} & \multicolumn{3}{c|}{\textbf{FIVR-200K}} & \multicolumn{3}{c|}{\textbf{CC\_WEB\_VIDEO}} & \multicolumn{3}{c|}{\textbf{SVD}} & \multicolumn{3}{c|}{\textbf{EVVE}} \\ \cline{3-14}
            & & \textbf{mAP} & \multicolumn{1}{c|}{\textbf{KB}} & \textbf{Sec} & \textbf{mAP} & \multicolumn{1}{c|}{\textbf{KB}} & \textbf{Sec} & \textbf{mAP} & \multicolumn{1}{c|}{\textbf{KB}} & \textbf{Sec} & \textbf{mAP} & \multicolumn{1}{c|}{\textbf{KB}} & \textbf{Sec} \\ \hline\hline
            \multirow{6}{*}{\rotatebox[origin=c]{90}{\centering \textbf{Coarse-grained}}}
            & \textbf{ITQ}~\citep{gong2012}                             & 0.491 & \textbf{0.1} & \textbf{0.733} & 0.954 & \textbf{0.1} & \textbf{0.045} & 0.842 & \textbf{0.1} & \textbf{1.793} & 0.606 & \textbf{0.1} & \textbf{0.005} \\ \cline{2-14}
            & \textbf{MFH}~\citep{song2013}                             & 0.525 & \textbf{0.1} & \textbf{0.733} & 0.967 & \textbf{0.1} & \textbf{0.045} & 0.849 & \textbf{0.1} & \textbf{1.793} & 0.604 & \textbf{0.1} & \textbf{0.005} \\ \cline{2-14}
            & \textbf{BoW}~\citep{cai2011}                              & \textbf{0.699} & 0.3 & 1.540 & 0.977 & 0.3 & 0.053 & 0.568 & \textbf{0.1} & 3.308 & 0.539 & 0.2 & \textbf{0.005} \\ \cline{2-14}
            & \textbf{DML}~\citep{kordopatis2017b}                      & 0.503 & 2    & 0.769 & 0.965 & 2    & 0.047 & 0.850 & 2    & 1.915 & 0.611 & 2    & 0.006 \\ \cline{2-14}
            & \textbf{TCA$_c$}~\citep{shao2021}                         & 0.570 & 4    & 0.812 & 0.965 & 4    & 0.047 &  -    & -         & -     & 0.598$\ssymbol{1}$ & 4    & 0.006 \\ \cline{2-14}
            & \textbf{Coarse-grained student - }$\textbf{S}^c$ (Ours)   & 0.574 & 4    & 0.812 & 0.967 & 4    & 0.047 & 0.868 & 4    & 1.920 & 0.636 & 4    & 0.006 \\ \hline\hline
            \multirow{6}{*}{\rotatebox[origin=c]{90}{\centering \textbf{Fine-grained}}}
            & \textbf{TMK}~\citep{poullot2015}                                           & 0.524 & 256   & 119.8 & 0.975 & 256   & 6.949 & 0.863 & 256  & 282.5   & 0.618 & 256   & 1.010  \\ \cline{2-14}
            & \textbf{LAMV}~\citep{baraldi2018}                                          & 0.619 & 256   & 167.2 & 0.975 & 256   & 9.703 & 0.880 & 256  & 394.5   & 0.620 & 256   & 1.410  \\ \cline{2-14}
            & \textbf{TCA$_f$}~\citep{shao2021}                                          & 0.877 & 438   & \textbf{36.15} & 0.990 & 596   & \textbf{2.097} &  -    & -    & -       & 0.603$\ssymbol{1}$ & 932 & \textbf{0.228} \\ \cline{2-14}
            & \textbf{ViSiL}~\citep{kordopatis2019b}                                     & 0.899 & 15124 & 451.9 & \textbf{0.992} & 20111 & 24.26 & 0.881 & 2308 & 319.8  & \textbf{0.658}  & 31457 & 5.718 \\ \cline{2-14}
            & \textbf{Fine-grained att. student - }$\textbf{S}^f_\mathcal{A}$ (Ours)     & \textbf{0.921} & 2016  & 149.1 & \textbf{0.992} & 2682  & 8.260 & \textbf{0.902} & 308  & 271.8  & 0.651  & 4194  & 1.506 \\ \cline{2-14}
            & \textbf{Fine-grained bin. student - }$\textbf{S}^f_\mathcal{B}$ (Ours)     & 0.909 & \textbf{63}    & 146.9 & 0.991 & \textbf{84}    & 8.129 & 0.891 & \textbf{10}   & \textbf{266.5}  & 0.640  & \textbf{131}   & 1.487 \\ \hline\hline
            \multirow{6}{*}{\rotatebox[origin=c]{90}{\centering \textbf{Re-ranking}}}
            & \textbf{TMK+QE}~\citep{poullot2015}     & 0.580 & 256  & 239.6 & 0.976 & 256  & 13.90 & 0.774 & 256  & 576.0 & 0.648 & 256  & 2.020 \\ \cline{2-14}
            & \textbf{LAMV+QE}~\citep{baraldi2018}    & 0.659 & 256  & 334.4 & \textbf{0.984} & 256  & 19.41 & 0.786 & 256  & 766.0 & \textbf{0.653} & 256  & 2.820 \\ \cline{2-14}
            & $\textbf{DnS}^{5\%}_\mathcal{A}$ (Ours) & 0.874 & 2020 & 8.267 & 0.969 & 2686 & 0.463 & 0.895 & 312  & 15.41 & 0.594 & 4198 & 0.081 \\ \cline{2-14}
            & $\textbf{DnS}^{5\%}_\mathcal{B}$ (Ours) & 0.862 & \textbf{67}   & \textbf{8.154} & 0.967 & \textbf{88}   & \textbf{0.456} & 0.884 & \textbf{14}   & \textbf{15.14} & 0.584 & \textbf{135}  & \textbf{0.080} \\ \cline{2-14}
            & $\textbf{DnS}^{30\%}_\mathcal{A}$ (Ours)   & \textbf{0.913} & 2020 & 45.55 & 0.974 & 2686 & 2.528 & \textbf{0.902} & 312  & 83.36 & 0.646 & 4198 & 0.458 \\ \cline{2-14}
            & $\textbf{DnS}^{30\%}_\mathcal{B}$ (Ours)          & 0.900 & \textbf{67}   & 44.87 & 0.974 & \textbf{88}   & 2.489 & 0.894 & \textbf{14}   & 81.76 & 0.634 & \textbf{135}  & 0.452 \\ \hline
          \end{tabular}
          }
          \label{tab:storage_time}
        \end{table*}
        
        In Table~\ref{tab:mAP_comparison}, the mAP of the proposed method in comparison to the video retrieval methods from the literature is reported. The proposed students achieve very competitive performance achieving state-of-the-art results in several cases. First, the fine-grained attention student achieves the best results on the two large-scale datasets, i.e., FIVR-200K and SVD, outperforming ViSiL (our teacher network) by a large margin, i.e., 0.022 and 0.021 mAP, respectively. It reports almost the same performance as ViSiL on the CC\_WEB\_VIDEO dataset, and it is slightly outperformed on the EVVE dataset. Additionally, it is noteworthy that the fine-grained binarization student demonstrates very competitive performance on all datasets. It achieves similar performance with ViSiL and the fine-grained attention student on the CC\_WEB\_VIDEO, the second-best results on all three tasks of FIVR-200K, and the third-best on SVD with a small margin from the second-best. However, its performance is lower than the teacher's on the EVVE dataset, highlighting that feature reduction and hashing have considerable impact on the student's retrieval performance on this dataset. Also, another possible explanation for this performance difference could be that the training dataset does not cover the included events sufficiently. 
        
        Second, the coarse-grained student exhibits very competitive performance among coarse-grained approaches on all datasets. It achieves the best mAP on two out of four evaluation datasets, i.e., on SVD and EVVE, reporting performance close or even better than several fine-grained methods. On FIVR-200K and CC\_WEB\_VIDEO, it is outperformed by the BoW-based approaches, which are trained with samples from the evaluation sets. However, when they are built with video corpora other than the evaluation (which simulates more realistic scenarios), their performance drops considerably~\citep{kordopatis2017b,kordopatis2019a}. Also, their performance on the SVD and EVVE datasets is considerably lower. 
        
        Third, our DnS runs maintain competitive performance. It improves the performance of the coarse-grained student by more than 0.2 on FIVR-200K and 0.02 on SVD by re-ranking only 5\% of the dataset with the fine-grained students. However, on the other two datasets, i.e., CC\_WEB\_VIDEO and EVVE, the re-ranking has negative effects on performance. A possible explanation for this might be that the performance of the coarse- and fine-grained students is very close, especially on the EVVE dataset. Also, this dataset consists of longer videos than the rest, which may impact the selection process. Nevertheless, the performance drop on these two datasets is mitigated when 30\% of the dataset is sent to the fine-grained students for re-ranking; while on the FIVR-200K and SVD, the DnS method reaches the performance of the corresponding fine-grained students, or it even outperforms them, i.e., $\textbf{DnS}^{30\%}_\mathcal{B}$ outperforms $\textbf{S}^f_\mathcal{B}$ on SVD dataset.
        
        Additionally, Table \ref{tab:storage_time} displays the storage and time requirements and the reference performance of the proposed method on each dataset. In comparison, we include the video retrieval methods that are implemented with the same features and run on GPU. For FIVR-200K and CC\_WEB\_VIDEO datasets, we display the DSVR and cc\_web$^*_c$ runs, respectively. We have excluded the TN and DP methods, as they have been implemented on CPU and their transfer to GPU is non-trivial. Also, the requirements of the TCA runs from \citep{shao2021} are approximated based on features of the same dimensionality. All times are measured on a Linux machine with the Intel i9-7900X CPU and an Nvidia 2080Ti GPU. 
        
        First, the individual students are compared against the competing methods in their corresponding category. The fine-grained binarization student has the lowest storage requirements among the fine-grained approaches on all datasets, having 240 times lower storage requirements than the ViSiL teacher. The fine-grained attention student needs the second-highest requirements in terms of space, but still, it needs 7.5 times less than ViSiL, achieving considerably better retrieval performance on two out of four evaluation datasets. However, the required retrieval time is high for all fine-grained approaches, especially in comparison with the coarse-grained ones. The coarse-grained student, which employs global vectors, has high storage requirements compared to the hashing and BoW methods that need notably lower storage space. In terms of time, all coarse-grained methods need approximately the same on all datasets, which is several orders of magnitude faster than the fine-grained ones. 
        
        Second, we benchmark our DnS approach with the two fine-grained students and two dataset percentages sent for refinement. An excellent trade-off between time and performance comes with the $\textbf{DnS}^{5\%}_\mathcal{B}$ offering an acceleration of more than 17 times in comparison to the fine-grained students, at a small cost in terms of performance when 5\% is used. Combined with the fine-grained binarization student, on FIVR-200K, it offers 55 times faster retrieval and 240 times lower storage requirements compared to the original ViSiL teacher providing comparable retrieval performance, i.e., 0.041 relative drop in terms of mAP. The performance of the DnS increases considerably when 30\% of the video pairs are sent for re-ranking, outperforming the ViSiL on two datasets with considerable margins. However, this performance improvement comes with a corresponding increase in the retrieval time.

%%%%%%%%%%%%%%%%%%%%%%%%%%%%%%%%%%%%%%%%%%%%%%%%%%%%%%%%%%%%%%%%%%%%%%%%%%%%%%%%
%%                                                                            %%
%%                              [Conclusion]                                  %%
%%                                                                            %%
%%%%%%%%%%%%%%%%%%%%%%%%%%%%%%%%%%%%%%%%%%%%%%%%%%%%%%%%%%%%%%%%%%%%%%%%%%%%%%%%
\section{Conclusion}\label{sec:conclusion}

    In this paper, we proposed a video retrieval framework based on Knowledge Distillation that addresses the problem of performance-efficiency trade-off focused on large-scale datasets. In contrast to typical video retrieval methods that rely on either a high-performance but resource demanding fine-grained approach or a computationally efficient but low-performance coarse-grained one, we introduced a Distill-and-Select approach. Several student networks were trained via a Teacher-Student setup at different performance-efficiency trade-offs. We experimented with two fine-grained students, one with a more elaborate attention mechanism that achieves better performance and one using a binarization layer offering very high performance with significantly lower storage requirements. Additionally, we trained a coarse-grained student that provides very fast retrieval with low storage requirements but at a high cost in performance. Once the students were trained, we combined them using a selector network that directs samples to the appropriate student in order to achieve high performance with high efficiency. It was trained based on the similarity difference between a coarse-grained and a fine-grained student so as to decide at query-time whether the similarity calculated by the coarse-grained one is reliable or the fine-grained one needs to be applied. The proposed method has been benchmarked to a number of content-based video retrieval datasets, where it improved the state-of-art in several cases and achieved very competitive performance with a remarkable reduction of the computational requirements.
    
    The proposed scheme can be employed with several setups based on the requirements of the application. For example, when small-scale databases are involved, with no strict storage space and computational time restrictions, the fine-grained attention student could be employed since it achieves the best retrieval performance. On the other hand, for mid-scale databases, where the storage requirements increase, the fine-grained binarization student would be a reasonable option since it achieves very high retrieval performance with remarkable reduction of storage space requirements. Finally, for large-scale databases, where both storage space and computation time are an issue, the combination of fine-grained binarization student and the coarse-grained student with the selector network would be an appropriate solution that offers high retrieval performance and high efficiency.
    
    % A possible limitation of our proposed approach could lie in the collected dataset. We followed the same protocol as in FIVR-200K, gathering videos from news events related to the categories of armed conflicts and natural disasters. This process generated a domain-focused video collection that does not cover more general cases, as in the case of the EVVE dataset. This could be a potential explanation for the performance difference between our proposed students compared to the teacher in this dataset since this is the only case where our students were outperformed. To this end, using a more general video dataset, i.e., using all news events to collect videos (not only the ones associated with the two categories), could lead to better retrieval performance.
    
    In the future, we plan to investigate alternatives for the better selection and re-ranking of video pairs based on our selector network by exploiting the ranking of videos derived from the two students. Also, we will explore better architectural choices for the development of the coarse-grained student to further improve the system's scalability with little compromises in retrieval performance.

\begin{acknowledgements}
This work has been supported by the projects MediaVerse and AI4Media, partially funded by the European Commission under contract number 957252 and 951911, respectively, and DECSTER funded by EPSRC under contract number EP/R025290/1.
\end{acknowledgements}

\balance

% Authors must disclose all relationships or interests that 
% could have direct or potential influence or impart bias on 
% the work: 
%
% \section*{Conflict of interest}
%
% The authors declare that they have no conflict of interest.

% BibTeX users please use one of
\bibliographystyle{spbasic}      % basic style, author-year citations
\bibliography{ref.bib}   % name your BibTeX data base

\begin{thebibliography}{79}
\providecommand{\natexlab}[1]{#1}
\providecommand{\url}[1]{{#1}}
\providecommand{\urlprefix}{URL }
\expandafter\ifx\csname urlstyle\endcsname\relax
  \providecommand{\doi}[1]{DOI~\discretionary{}{}{}#1}\else
  \providecommand{\doi}{DOI~\discretionary{}{}{}\begingroup
  \urlstyle{rm}\Url}\fi
\providecommand{\eprint}[2][]{\url{#2}}

\bibitem[{Arandjelovic et~al.(2016)Arandjelovic, Gronat, Torii, Pajdla, and
  Sivic}]{arandjelovic2016}
Arandjelovic R, Gronat P, Torii A, Pajdla T, Sivic J (2016) {NetVLAD: CNN
  architecture for weakly supervised place recognition}. In: Proceedings of the
  IEEE conference on Computer Vision and Pattern Recognition

\bibitem[{Ba et~al.(2016)Ba, Kiros, and Hinton}]{ba2016}
Ba JL, Kiros JR, Hinton GE (2016) Layer normalization. arXiv preprint
  arXiv:160706450

\bibitem[{Baraldi et~al.(2018)Baraldi, Douze, Cucchiara, and
  J{\'e}gou}]{baraldi2018}
Baraldi L, Douze M, Cucchiara R, J{\'e}gou H (2018) {LAMV}: Learning to align
  and match videos with kernelized temporal layers. In: Proceedings of the IEEE
  conference on Computer Vision and Pattern Recognition

\bibitem[{Bhardwaj et~al.(2019)Bhardwaj, Srinivasan, and Khapra}]{bhardwaj2019}
Bhardwaj S, Srinivasan M, Khapra MM (2019) Efficient video classification using
  fewer frames. In: Proceedings of the IEEE Conference on Computer Vision and
  Pattern Recognition

\bibitem[{Bishay et~al.(2019)Bishay, Zoumpourlis, and Patras}]{bishay2019}
Bishay M, Zoumpourlis G, Patras I (2019) {TARN}: Temporal attentive relation
  network for few-shot and zero-shot action recognition. In: Proceedings of the
  British Machine Vision Conference

\bibitem[{Cai et~al.(2011)Cai, Yang, Ping, Wang, Mei, Hua, and Li}]{cai2011}
Cai Y, Yang L, Ping W, Wang F, Mei T, Hua XS, Li S (2011) Million-scale
  near-duplicate video retrieval system. In: Proceedings of the ACM
  international conference on Multimedia, ACM

\bibitem[{Cao et~al.(2017)Cao, Long, Wang, and Yu}]{cao2017}
Cao Z, Long M, Wang J, Yu PS (2017) {HashNet}: Deep learning to hash by
  continuation. In: Proceedings of the IEEE International Conference on
  Computer Vision

\bibitem[{Chou et~al.(2015)Chou, Chen, and Lee}]{chou2015}
Chou CL, Chen HT, Lee SY (2015) Pattern-based near-duplicate video retrieval
  and localization on web-scale videos. IEEE Transactions on Multimedia
  17(3):382--395

\bibitem[{Chum et~al.(2007)Chum, Philbin, Sivic, Isard, and
  Zisserman}]{chum2007}
Chum O, Philbin J, Sivic J, Isard M, Zisserman A (2007) Total recall: Automatic
  query expansion with a generative feature model for object retrieval. In:
  Proceedings of the IEEE International Conference on Computer Vision

\bibitem[{Crasto et~al.(2019)Crasto, Weinzaepfel, Alahari, and
  Schmid}]{crasto2019}
Crasto N, Weinzaepfel P, Alahari K, Schmid C (2019) Mars: Motion-augmented rgb
  stream for action recognition. In: Proceedings of the IEEE Conference on
  Computer Vision and Pattern Recognition

\bibitem[{Deng et~al.(2019)Deng, Pan, Yao, Zhou, Li, and Mei}]{deng2019}
Deng J, Pan Y, Yao T, Zhou W, Li H, Mei T (2019) Relation distillation networks
  for video object detection. In: Proceedings of the IEEE International
  Conference on Computer Vision

\bibitem[{Douze et~al.(2010)Douze, J{\'e}gou, and Schmid}]{douze2010}
Douze M, J{\'e}gou H, Schmid C (2010) An image-based approach to video copy
  detection with spatio-temporal post-filtering. IEEE Transactions on
  Multimedia 12(4):257--266

\bibitem[{Douze et~al.(2013)Douze, Revaud, Schmid, and J{\'e}gou}]{douze2013}
Douze M, Revaud J, Schmid C, J{\'e}gou H (2013) Stable hyper-pooling and query
  expansion for event detection. In: Proceedings of the IEEE International
  Conference on Computer Vision, pp 1825--1832

\bibitem[{Feng et~al.(2018)Feng, Ma, Liu, Zhang, and Luo}]{feng2018}
Feng Y, Ma L, Liu W, Zhang T, Luo J (2018) Video re-localization. In:
  Proceedings of the European Conference on Computer Vision

\bibitem[{Gao et~al.(2017)Gao, Hua, Zhang, Jojic, Wang, Xue, and
  Zheng}]{gao2017}
Gao Z, Hua G, Zhang D, Jojic N, Wang L, Xue J, Zheng N (2017) {ER3}: A unified
  framework for event retrieval, recognition and recounting. In: Proceedings of
  the IEEE conference on Computer Vision and Pattern Recognition

\bibitem[{Garcia et~al.(2018)Garcia, Morerio, and Murino}]{garcia2018}
Garcia NC, Morerio P, Murino V (2018) Modality distillation with multiple
  stream networks for action recognition. In: Proceedings of the European
  Conference on Computer Vision

\bibitem[{Gong et~al.(2012)Gong, Lazebnik, Gordo, and Perronnin}]{gong2012}
Gong Y, Lazebnik S, Gordo A, Perronnin F (2012) Iterative quantization: A
  procrustean approach to learning binary codes for large-scale image
  retrieval. IEEE Transactions on Pattern Analysis and Machine Intelligence
  35(12):2916--2929

\bibitem[{Gordo et~al.(2020)Gordo, Radenovic, and Berg}]{gordo2020}
Gordo A, Radenovic F, Berg T (2020) Attention-based query expansion learning.
  In: Proceedings of the European Conference on Computer Vision

\bibitem[{Gou et~al.(2021)Gou, Yu, Maybank, and Tao}]{gou2021}
Gou J, Yu B, Maybank SJ, Tao D (2021) Knowledge distillation: A survey.
  International Journal of Computer Vision pp 1--31

\bibitem[{He et~al.(2016)He, Zhang, Ren, and Sun}]{he2016}
He K, Zhang X, Ren S, Sun J (2016) Deep residual learning for image
  recognition. In: Proceedings of the IEEE Conference on Computer Vision and
  Pattern Recognition

\bibitem[{Hinton et~al.(2015)Hinton, Vinyals, and Dean}]{hinton2015}
Hinton G, Vinyals O, Dean J (2015) Distilling the knowledge in a neural
  network. In: Proceedings of the International Conference on Neural
  Information Processing Systems

\bibitem[{Huang et~al.(2010)Huang, Shen, Shao, Cui, and Zhou}]{huang2010}
Huang Z, Shen HT, Shao J, Cui B, Zhou X (2010) Practical online near-duplicate
  subsequence detection for continuous video streams. IEEE Transactions on
  Multimedia 12(5):386--398

\bibitem[{Ioffe and Szegedy(2015)}]{ioffe2015}
Ioffe S, Szegedy C (2015) Batch normalization: Accelerating deep network
  training by reducing internal covariate shift. In: Proceedings of the
  International Conference on Machine Learning

\bibitem[{J{\'e}gou and Chum(2012)}]{jegou2012}
J{\'e}gou H, Chum O (2012) Negative evidences and co-occurences in image
  retrieval: The benefit of pca and whitening. In: Proceedings of the European
  Conference on Computer Vision, Springer, pp 774--787

\bibitem[{Jiang et~al.(2019)Jiang, He, Li, Lin, Li, and Li}]{jiang2019}
Jiang QY, He Y, Li G, Lin J, Li L, Li WJ (2019) {SVD}: A large-scale short
  video dataset for near-duplicate video retrieval. In: Proceedings of the IEEE
  International Conference on Computer Vision

\bibitem[{Jiang and Wang(2016)}]{jiang2016}
Jiang YG, Wang J (2016) Partial copy detection in videos: A benchmark and an
  evaluation of popular methods. IEEE Trans Big Data 2(1):32--42

\bibitem[{Jiang et~al.(2014)Jiang, Jiang, and Wang}]{jiang2014}
Jiang YG, Jiang Y, Wang J (2014) {VCDB}: A large-scale database for partial
  copy detection in videos. In: Proceedings of the European Conference on
  Computer Vision, Springer, pp 357--371

\bibitem[{Kingma and Ba(2015)}]{kingma2014}
Kingma D, Ba J (2015) Adam: A method for stochastic optimization. In:
  Proceedings of the International Conference on Learning Representations

\bibitem[{Kordopatis-Zilos et~al.(2017{\natexlab{a}})Kordopatis-Zilos,
  Papadopoulos, Patras, and Kompatsiaris}]{kordopatis2017a}
Kordopatis-Zilos G, Papadopoulos S, Patras I, Kompatsiaris I
  (2017{\natexlab{a}}) Near-duplicate video retrieval by aggregating
  intermediate cnn layers. In: Proceedings of the International Conference on
  Multimedia Modeling, Springer, pp 251--263

\bibitem[{Kordopatis-Zilos et~al.(2017{\natexlab{b}})Kordopatis-Zilos,
  Papadopoulos, Patras, and Kompatsiaris}]{kordopatis2017b}
Kordopatis-Zilos G, Papadopoulos S, Patras I, Kompatsiaris I
  (2017{\natexlab{b}}) Near-duplicate video retrieval with deep metric
  learning. In: Proceedings of the IEEE International Conference on Computer
  Vision Workshops, IEEE, pp 347--356

\bibitem[{Kordopatis-Zilos et~al.(2019{\natexlab{a}})Kordopatis-Zilos,
  Papadopoulos, Patras, and Kompatsiaris}]{kordopatis2019a}
Kordopatis-Zilos G, Papadopoulos S, Patras I, Kompatsiaris I
  (2019{\natexlab{a}}) {FIVR}: Fine-grained incident video retrieval. IEEE
  Transactions on Multimedia

\bibitem[{Kordopatis-Zilos et~al.(2019{\natexlab{b}})Kordopatis-Zilos,
  Papadopoulos, Patras, and Kompatsiaris}]{kordopatis2019b}
Kordopatis-Zilos G, Papadopoulos S, Patras I, Kompatsiaris I
  (2019{\natexlab{b}}) {ViSiL}: Fine-grained spatio-temporal video similarity
  learning. In: Proceedings of the IEEE International Conference on Computer
  Vision

\bibitem[{Krizhevsky et~al.(2012)Krizhevsky, Sutskever, and
  Hinton}]{krizhevsky2012}
Krizhevsky A, Sutskever I, Hinton GE (2012) Imagenet classification with deep
  convolutional neural networks. In: Proceedings of the International
  Conference on Neural Information Processing Systems

\bibitem[{Lassance et~al.(2020)Lassance, Bontonou, Hacene, Gripon, Tang, and
  Ortega}]{lassance2020}
Lassance C, Bontonou M, Hacene GB, Gripon V, Tang J, Ortega A (2020) Deep
  geometric knowledge distillation with graphs. In: Proceedings of the IEEE
  International Conference on Acoustics, Speech and Signal Processing

\bibitem[{Lee et~al.(2020)Lee, Lee, Ng, and Natsev}]{lee2020}
Lee H, Lee J, Ng JYH, Natsev P (2020) Large scale video representation learning
  via relational graph clustering. In: Proceedings of the IEEE conference on
  Computer Vision and Pattern Recognition

\bibitem[{Lee et~al.(2018)Lee, Abu-El-Haija, Varadarajan, and Natsev}]{lee2018}
Lee J, Abu-El-Haija S, Varadarajan B, Natsev A (2018) Collaborative deep metric
  learning for video understanding. In: Proceedings of the ACM SIGKDD
  International Conference on Knowledge Discovery \& Data Mining

\bibitem[{Li et~al.(2017)Li, Jin, and Yan}]{li2017}
Li Q, Jin S, Yan J (2017) Mimicking very efficient network for object
  detection. In: Proceedings of the IEEE Conference on Computer Vision and
  Pattern Recognition

\bibitem[{Liang et~al.(2019)Liang, Lin, Wang, Shao, Wang, and Chen}]{liang2019}
Liang D, Lin L, Wang R, Shao J, Wang C, Chen YW (2019) Unsupervised
  teacher-student model for large-scale video retrieval. In: Proceedings of the
  IEEE International Conference on Computer Vision Workshops

\bibitem[{Liang and Wang(2020)}]{liang2020}
Liang S, Wang P (2020) An efficient hierarchical near-duplicate video detection
  algorithm based on deep semantic features. In: Proceedings of the
  International Conference on Multimedia Modeling

\bibitem[{Liao et~al.(2018)Liao, Lei, Zheng, Lin, Cao, Zhang, and
  Ding}]{liao2018}
Liao K, Lei H, Zheng Y, Lin G, Cao C, Zhang M, Ding J (2018) {IR} feature
  embedded bof indexing method for near-duplicate video retrieval. IEEE
  Transactions on Circuits and Systems for Video Technology 29(12):3743--3753

\bibitem[{Liong et~al.(2017)Liong, Lu, Tan, and Zhou}]{liong2017}
Liong VE, Lu J, Tan YP, Zhou J (2017) Deep video hashing. IEEE Transactions on
  Multimedia 19(6):1209--1219

\bibitem[{Liu et~al.(2017)Liu, Zhao, Wang, Lv, and Chen}]{liu2017}
Liu H, Zhao Q, Wang H, Lv P, Chen Y (2017) An image-based near-duplicate video
  retrieval and localization using improved edit distance. Multimedia Tools and
  Applications 76(22):24435--24456

\bibitem[{Liu et~al.(2019)Liu, Cao, Li, Yuan, Hu, Li, and Duan}]{liu2019}
Liu Y, Cao J, Li B, Yuan C, Hu W, Li Y, Duan Y (2019) Knowledge distillation
  via instance relationship graph. In: Proceedings of the IEEE Conference on
  Computer Vision and Pattern Recognition

\bibitem[{Luo et~al.(2018)Luo, Hsieh, Jiang, Niebles, and Fei-Fei}]{luo2018}
Luo Z, Hsieh JT, Jiang L, Niebles JC, Fei-Fei L (2018) Graph distillation for
  action detection with privileged modalities. In: Proceedings of the European
  Conference on Computer Vision

\bibitem[{Markatopoulou et~al.(2017)Markatopoulou, Galanopoulos, Mezaris, and
  Patras}]{markatopoulou2017}
Markatopoulou F, Galanopoulos D, Mezaris V, Patras I (2017) Query and keyframe
  representations for ad-hoc video search. In: Proceedings of the 2017 ACM on
  International Conference on Multimedia Retrieval, pp 407--411

\bibitem[{Markatopoulou et~al.(2018)Markatopoulou, Mezaris, and
  Patras}]{markatopoulou2018}
Markatopoulou F, Mezaris V, Patras I (2018) Implicit and explicit concept
  relations in deep neural networks for multi-label video/image annotation.
  IEEE transactions on circuits and systems for video technology
  29(6):1631--1644

\bibitem[{Miech et~al.(2017)Miech, Laptev, and Sivic}]{miech2017}
Miech A, Laptev I, Sivic J (2017) Learnable pooling with context gating for
  video classification. arXiv preprint arXiv:170606905

\bibitem[{Pan et~al.(2020)Pan, Cai, Huang, Lee, Gaidon, Adeli, and
  Niebles}]{pan2020}
Pan B, Cai H, Huang DA, Lee KH, Gaidon A, Adeli E, Niebles JC (2020)
  Spatio-temporal graph for video captioning with knowledge distillation. In:
  Proceedings of the IEEE Conference on Computer Vision and Pattern Recognition

\bibitem[{Park et~al.(2019)Park, Kim, Lu, and Cho}]{park2019}
Park W, Kim D, Lu Y, Cho M (2019) Relational knowledge distillation. In:
  Proceedings of the IEEE Conference on Computer Vision and Pattern Recognition

\bibitem[{Paszke et~al.(2019)Paszke, Gross, Massa, Lerer, Bradbury, Chanan,
  Killeen, Lin, Gimelshein, Antiga et~al.}]{paszke2019}
Paszke A, Gross S, Massa F, Lerer A, Bradbury J, Chanan G, Killeen T, Lin Z,
  Gimelshein N, Antiga L, et~al. (2019) {PyTorch}: An imperative style,
  high-performance deep learning library. In: Proceedings of the International
  Conference on Neural Information Processing Systems

\bibitem[{Peng et~al.(2019)Peng, Jin, Liu, Li, Wu, Liu, Zhou, and
  Zhang}]{peng2019}
Peng B, Jin X, Liu J, Li D, Wu Y, Liu Y, Zhou S, Zhang Z (2019) Correlation
  congruence for knowledge distillation. In: Proceedings of the IEEE
  International Conference on Computer Vision

\bibitem[{Piergiovanni et~al.(2020)Piergiovanni, Angelova, and
  Ryoo}]{piergiovanni2020}
Piergiovanni A, Angelova A, Ryoo MS (2020) Evolving losses for unsupervised
  video representation learning. In: Proceedings of the IEEE Conference on
  Computer Vision and Pattern Recognition

\bibitem[{Poullot et~al.(2015)Poullot, Tsukatani, Phuong~Nguyen, J{\'e}gou, and
  Satoh}]{poullot2015}
Poullot S, Tsukatani S, Phuong~Nguyen A, J{\'e}gou H, Satoh S (2015) Temporal
  matching kernel with explicit feature maps. In: Proceedings of the ACM
  international conference on Multimedia

\bibitem[{Revaud et~al.(2013)Revaud, Douze, Schmid, and J{\'e}gou}]{revaud2013}
Revaud J, Douze M, Schmid C, J{\'e}gou H (2013) Event retrieval in large video
  collections with circulant temporal encoding. In: Proceedings of the IEEE
  Conference on Computer Vision and Pattern Recognition, IEEE, pp 2459--2466

\bibitem[{Shao et~al.(2021)Shao, Wen, Zhao, and Xue}]{shao2021}
Shao J, Wen X, Zhao B, Xue X (2021) Temporal context aggregation for video
  retrieval with contrastive learning. In: Proceedings of the IEEE Winter
  Conference on Applications of Computer Vision

\bibitem[{Shmelkov et~al.(2017)Shmelkov, Schmid, and Alahari}]{shmelkov2017}
Shmelkov K, Schmid C, Alahari K (2017) Incremental learning of object detectors
  without catastrophic forgetting. In: Proceedings of the IEEE International
  Conference on Computer Vision

\bibitem[{Sivic and Zisserman(2003)}]{sivic2003}
Sivic J, Zisserman A (2003) {Video Google}: A text retrieval approach to object
  matching in videos. In: Proceedings of the IEEE Conference on Computer Vision
  and Pattern Recognition

\bibitem[{Song et~al.(2011)Song, Yang, Huang, Shen, and Hong}]{song2011}
Song J, Yang Y, Huang Z, Shen HT, Hong R (2011) Multiple feature hashing for
  real-time large scale near-duplicate video retrieval. In: Proceedings of the
  19th ACM international conference on Multimedia

\bibitem[{Song et~al.(2013)Song, Yang, Huang, Shen, and Luo}]{song2013}
Song J, Yang Y, Huang Z, Shen HT, Luo J (2013) Effective multiple feature
  hashing for large-scale near-duplicate video retrieval. IEEE Transactions on
  Multimedia 15(8):1997--2008

\bibitem[{Song et~al.(2018)Song, Zhang, Li, Gao, Wang, and Hong}]{song2018}
Song J, Zhang H, Li X, Gao L, Wang M, Hong R (2018) Self-supervised video
  hashing with hierarchical binary auto-encoder. IEEE Transactions on Image
  Processing 27(7):3210--3221

\bibitem[{Stroud et~al.(2020)Stroud, Ross, Sun, Deng, and
  Sukthankar}]{stroud2020}
Stroud J, Ross D, Sun C, Deng J, Sukthankar R (2020) D3d: Distilled 3d networks
  for video action recognition. In: Proceedings of the IEEE Winter Conference
  on Applications of Computer Vision

\bibitem[{Tan et~al.(2009)Tan, Ngo, Hong, and Chua}]{tan2009}
Tan HK, Ngo CW, Hong R, Chua TS (2009) Scalable detection of partial
  near-duplicate videos by visual-temporal consistency. In: Proceedings of the
  ACM international conference on Multimedia

\bibitem[{Tavakolian et~al.(2019)Tavakolian, Tavakoli, and
  Hadid}]{tavakolian2019}
Tavakolian M, Tavakoli HR, Hadid A (2019) {AWSD}: Adaptive weighted
  spatiotemporal distillation for video representation. In: Proceedings of the
  IEEE International Conference on Computer Vision

\bibitem[{Thoker and Gall(2019)}]{thoker2019}
Thoker FM, Gall J (2019) Cross-modal knowledge distillation for action
  recognition. In: Proceedings of the IEEE International Conference on Image
  Processing

\bibitem[{Tolias et~al.(2016)Tolias, Sicre, and J{\'e}gou}]{tolias2016}
Tolias G, Sicre R, J{\'e}gou H (2016) Particular object retrieval with integral
  max-pooling of cnn activations. In: Proceedings of the International
  Conference on Learning Representations

\bibitem[{Touvron et~al.(2020)Touvron, Cord, Douze, Massa, Sablayrolles, and
  J{\'e}gou}]{touvron2020}
Touvron H, Cord M, Douze M, Massa F, Sablayrolles A, J{\'e}gou H (2020)
  Training data-efficient image transformers \& distillation through attention.
  arXiv preprint arXiv:201212877

\bibitem[{Tung and Mori(2019)}]{tung2019}
Tung F, Mori G (2019) Similarity-preserving knowledge distillation. In:
  Proceedings of the IEEE International Conference on Computer Vision

\bibitem[{Vaswani et~al.(2017)Vaswani, Shazeer, Parmar, Uszkoreit, Jones,
  Gomez, Kaiser, and Polosukhin}]{vaswani2017}
Vaswani A, Shazeer N, Parmar N, Uszkoreit J, Jones L, Gomez AN, Kaiser L,
  Polosukhin I (2017) Attention is all you need. In: Proceedings of the
  International Conference on Neural Information Processing Systems

\bibitem[{Wang et~al.(2021)Wang, Cheng, Chen, Song, and Lai}]{wang2021}
Wang KH, Cheng CC, Chen YL, Song Y, Lai SH (2021) Attention-based deep metric
  learning for near-duplicate video retrieval. In: Proceedings of the
  International Conference on Pattern Recognition

\bibitem[{Wang et~al.(2017)Wang, Bao, Li, Fan, and Luo}]{wang2017}
Wang L, Bao Y, Li H, Fan X, Luo Z (2017) Compact cnn based video representation
  for efficient video copy detection. In: Proceedings of the International
  Conference on Multimedia Modeling

\bibitem[{Wu et~al.(2007)Wu, Hauptmann, and Ngo}]{wu2007}
Wu X, Hauptmann AG, Ngo CW (2007) Practical elimination of near-duplicates from
  web video search. In: Proceedings of the ACM international conference on
  Multimedia

\bibitem[{Xie et~al.(2020)Xie, Luong, Hovy, and Le}]{xie2020}
Xie Q, Luong MT, Hovy E, Le QV (2020) Self-training with noisy student improves
  imagenet classification. In: Proceedings of the IEEE Conference on Computer
  Vision and Pattern Recognition

\bibitem[{Yalniz et~al.(2019)Yalniz, J{'e}gou, Chen, Paluri, and
  Mahajan}]{yalniz2019}
Yalniz IZ, J{'e}gou H, Chen K, Paluri M, Mahajan D (2019) Billion-scale
  semi-supervised learning for image classification. arXiv preprint
  arXiv:190500546

\bibitem[{Yang et~al.(2019)Yang, Tian, and Huang}]{yang2019}
Yang Y, Tian Y, Huang T (2019) Multiscale video sequence matching for
  near-duplicate detection and retrieval. Multimedia Tools and Applications
  78(1):311--336

\bibitem[{Yang et~al.(2016)Yang, Yang, Dyer, He, Smola, and Hovy}]{yang2016}
Yang Z, Yang D, Dyer C, He X, Smola A, Hovy E (2016) Hierarchical attention
  networks for document classification. In: Proceedings of the Conference of
  the North American Chapter of the Association for Computational Linguistics:
  Human Language Technologies

\bibitem[{Yuan et~al.(2020)Yuan, Wang, Zhang, Tay, Jie, Liu, and
  Feng}]{yuan2020}
Yuan L, Wang T, Zhang X, Tay FE, Jie Z, Liu W, Feng J (2020) Central similarity
  quantization for efficient image and video retrieval. In: Proceedings of the
  IEEE conference on Computer Vision and Pattern Recognition

\bibitem[{Zhang and Peng(2018)}]{zhang2018}
Zhang C, Peng Y (2018) Better and faster: knowledge transfer from multiple
  self-supervised learning tasks via graph distillation for video
  classification. In: Proceedings of the International Joint Conference on
  Artificial Intelligence

\bibitem[{Zhang et~al.(2020)Zhang, Shi, Yuan, Li, Wang, Hu, and
  Zha}]{zhang2020}
Zhang Z, Shi Y, Yuan C, Li B, Wang P, Hu W, Zha ZJ (2020) Object relational
  graph with teacher-recommended learning for video captioning. In: Proceedings
  of the IEEE conference on computer vision and pattern recognition

\bibitem[{Zhao et~al.(2019)Zhao, Chen, Chen, Li, Xiang, Zhao, and
  Su}]{zhao2019}
Zhao Z, Chen G, Chen C, Li X, Xiang X, Zhao Y, Su F (2019) Instance-based video
  search via multi-task retrieval and re-ranking. In: Proceedings of the
  IEEE/CVF International Conference on Computer Vision Workshops, pp 0--0

\end{thebibliography}

\end{document}